%% file: acl_latex.tex
\documentclass[11pt]{article}

\usepackage[preprint]{acl}

\usepackage{times}
\usepackage{latexsym}
\usepackage{amssymb}
\usepackage{amsmath}
\usepackage{subcaption}
\usepackage[T1]{fontenc}

\usepackage[utf8]{inputenc}

\usepackage{microtype}

\usepackage{inconsolata}

\usepackage{graphicx}
\usepackage{tabularx}
\usepackage{spverbatim}
\usepackage{enumitem}
\usepackage[most]{tcolorbox}
\usepackage{float}
\usepackage{booktabs}
\usepackage{multirow}
\usepackage{tabularx}
\usepackage{geometry} 
\newcolumntype{Y}{>{\centering\arraybackslash}X}
\newtcolorbox{promptbox}[1][]{
  colback=gray!5!white, 
  colframe=gray!75!black, 
  fonttitle=\bfseries,
  title={#1}, 
  arc=0mm, 
  boxrule=0.5pt,
  left=5pt, right=5pt, top=5pt, bottom=5pt
}

\title{The GDN-CC Dataset: Automatic Corpus Clarification for AI-enhanced Democratic Citizen Consultations}

\author{
 \textbf{Pierre-Antoine Lequeu\textsuperscript{1}}, \
 \textbf{Léo Labat\textsuperscript{1,3}}, \ 
 \textbf{Laurène Cave\textsuperscript{2}}, \ 
 \textbf{Gaël Lejeune\textsuperscript{2}},
\\
 \textbf{François Yvon\textsuperscript{1}}
 \and \textbf{Benjamin Piwowarski\textsuperscript{1}}
\\
 \textsuperscript{1} Sorbonne Université, CNRS, ISIR, Paris, France \\
 \textsuperscript{2}  Sorbonne Université, STIH/CERES, Paris, France \\
 \textsuperscript{3}  Institut Polytechnique de Paris, CNRS, CREST, Paris, France
\\
 \small{
   \textbf{Correspondence:} \href{mailto:lequeu@isir.upmc.fr}{lequeu \textit{(at)} isir.upmc.fr}
 }
}

\begin{document}
\maketitle
\begin{abstract}
LLMs are ubiquitous in modern NLP, and while their applicability extends to texts produced for democratic activities such as online deliberations or large-scale citizen consultations, ethical questions have been raised for their usage as analysis tools. We continue this line of research with two main goals: (a) to develop resources that can help standardize citizen contributions in public forums at the \textbf{pragmatic level}, and make them easier to use in topic modeling and political analysis; (b) to study how well this standardization can reliably be performed by small,  open-weights LLMs, \emph{i.e.} models that can be run locally and transparently with limited resources. Accordingly, we introduce \textbf{Corpus Clarification} as a preprocessing framework for large-scale consultation data that transforms noisy, multi-topic contributions into structured, self-contained argumentative units ready for downstream analysis. We present \textbf{GDN-CC}, a manually-curated dataset of 1,231 contributions to the French \textit{Grand Débat National}, comprising 2,285 argumentative units annotated for argumentative structure and manually clarified. 
We then show that finetuned Small Language Models match or outperform LLMs on reproducing these annotations, and measure their usability for an opinion clustering task. We finally release \textbf{GDN-CC-large}, an automatically annotated corpus of 240k contributions, the largest annotated democratic consultation dataset to date.
\end{abstract}

\begin{figure*}
    \centering
\includegraphics[width=0.99\linewidth]{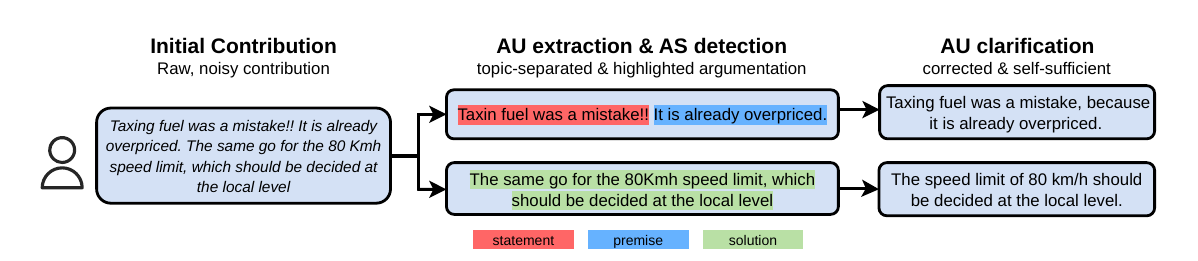}
    \caption{The Corpus Clarification task applied to one contribution. The original contribution (left) is segmented in topics, then in argumentative units (middle), which are automatically reformulated to generate clearer versions of the expressed opinions (right). The contribution was translated to English and simplified for illustrative purposes.}
    \label{fig:explaination}
\end{figure*}

\section{Introduction}
\label{part:introduction}
\input{parts/introduction}

\section{Related Work}
\label{part:related_work}
\input{parts/related_works}

\section{The Corpus Clarification task}
\label{part:task}
\input{parts/corpusclarif}

\section{Annotation Process}
\label{part:annotation}
\input{parts/annotation_process}

\section{GDN-CC: A French Dataset for the task of Corpus Clarification}
\label{part:corpus}
\input{parts/dataset}
\section{Experiments}
\label{part:experiments}

\input{parts/experiments}

\begin{table*}[t]
  \centering
  \begin{tabular}{lc|c|ccc}
    \hline
    \textbf{Theme} & \textbf{Contribs.} & \textbf{AUs}  & \textbf{statements} & \textbf{solutions} & \textbf{premises} \\
    \hline
    \textbf{Taxation and Public Spending} & 95,689 & 120,902 & 56,709\tiny{(28.2\%)} & 118,399\tiny{(59.0\%)} & 25,666\tiny{(12.8\%)} \\
    \textbf{Ecological Transition} & 77,944 & 100,442 & 50,872\tiny{(31.3\%)} & 92,587\tiny{(56.9\%)} & 19,182\tiny{(11.8\%)} \\
    \textbf{Organization of the State} & 28,584 & 35,204 & 19,948\tiny{(32.6\%)} & 32,826\tiny{(53.6\%)} & 8,418\tiny{(13.8\%)}\\
    \textbf{Democracy and Citizenship} & 37,581 & 44,200 & 28,217\tiny{(37.7\%)} & 37,854\tiny{(50.6\%)} & 8,771\tiny{(11.7\%)} \\
    \hline
    \textbf{Total}  & \textbf{239,798} & \textbf{300,748} & \textbf{155,746}\tiny{(31.2\%)} & \textbf{281,666}\tiny{(56.4\%)} & \textbf{62,037}\tiny{(12.4\%)}\\
    \hline
  \end{tabular}
  \caption{
    Number of contributions, argumentative units, statements, solutions, and premises per theme and in total in \textbf{GDN-CC-large}, the automatically-annotated corpus. Percentages are calculated row-wise.
  }
   \label{tab:stats_final_corpus}
\end{table*}

\input{parts/downstream}

\input{parts/full_corpus}

\section{Conclusion}
In this work, we formalized \textbf{Corpus Clarification} as a first step for the automated analysis of large-scale democratic consultations. By defining and evaluating a three-step pipeline, we have developed a structured methodology to transform noisy, multi-topic citizen contributions into standardized, self-sufficient opinions. Our results, based on the manual annotation of \textbf{GDN-CC} demonstrate that finetuned encoders or SLMs match or outperform large proprietary models. Specifically, derivatives from \texttt{Qwen2.5-7b} and \texttt{Gemma-2-9b} proved amply sufficient to obtain high quality annotations. 
Furthermore, we provided empirical evidence that this clarification process significantly enhances the performance of embedding-based clustering, a critical prerequisite for reliable topic modeling and debate facilitation at scale. By releasing the 240k processed contributions \textbf{GDN-CC-large}, we provide the research community with the largest democratic consultation corpus annotated for downstream tasks.

\section*{Limitations}
Though we bring a new approach to LLMs for citizen consultation analysis, we acknowledge some limitations to our work. While the Corpus Clarification task was thought of as language and topic-agnostic, our work only focus on French contributions for the \textit{Grand Débat National} and do not explore other languages or other types of consultations, with potential unforeseen language- or consultation-specific limitations to this framework. 
Moreover, we did not perform any extensive exploration of LLM's prompts, which could further improve their capacities for the Corpus Clarification task. We also focus on a restricted set of LLMs systems (both for the manual annotation and the automatic annotation). Considering a larger set could bring deeper insights about the capacities of various systems for this task. 
Finally, some evaluations rely on LLM-as-a-Judge which, while giving statistically significant results, is known to be sensitive to prompting.

\section*{Ethical Considerations}
The deployment of AI in democratic processes necessitates a rigorous examination of the trade-offs between computational efficiency and representational fidelity. The argumentative unit clarification step of our pipeline prioritizes legibility to facilitate large-scale analysis. However, standardizing raw expression carries the inherent risk of discarding sentiment, nuance, or the unique "voice" of the contributor. While we demonstrate that this process significantly improves downstream clustering, an ideal democratic analysis should not completely forget the original form of expression. Moreover, while finetuned LLMs reduce reliance on opaque systems for analysis, the models we used are still developed by private entities. This does not fully resolve the problem of private dependence in public democratic infrastructure.

Concerning the data from the \textit{Grand Débat National} that we use, all participants were aware that their answers would be shared publicly. While anonymity and the removal of hate speech was ensured for the manual annotation, we did not further check if there was potential identifying information or hate speech in the automatically-annotated data. However, the original data source from the French open data platform affirms that the data has been anonymized.


\section*{Acknowledgments}
This research was funded by BPI-France under the project AI For Democracy - Democratic Commons, one of seven winners of BPI-France’s “Digital Commons for Generative AI” call for projects, conducted as part of the France 2030 investment plan. We thank the annotators for their work as well as all members of the Democratic Commons project, in particular Paul Lerner and Nazanin Shafiabadi. This work was performed using HPC resources from GENCI–IDRIS (Grant 2024-AD011015927).


\bibliography{custom,anthology-1,anthology-2}


\appendix
\input{parts/Appendix}
\end{document}

%% file: parts/introduction.tex
The huge improvement of Natural Language Processing (NLP) systems in the last decade, owing to the rapid adoption of the transformer architecture \cite{vaswani2017attention}, has made Large Language Models (LLMs) a key component for processing texts in fields such as healthcare \cite{nazi2024large}, education \cite{yan2024practical}, or business and industry decision-making \cite{chkirbene2024large}.
Applications of LLMs in the context of politics and democratic activities have also quickly been considered \cite{aoki2024large,coeckelbergh2025llms, summerfield2024will}. However, using such powerful tools in end-to-end systems for political analysis and decision-making raises serious ethical questions \cite{galariotisArtificialIntelligenceThreatening2024, revelAIFacilitatedCollectiveJudgements2025}. 
This is due to the opacity of the decision-making process and the difficulty to steer it through human intervention. Explanations delivered by Chain-of-Thought methods hardly mitigate these defects, as discussed, \textit{inter alia}, by \citet{barez2025chain}.
LLMs outputs also tend to be heavily biased along multiple dimensions such as gender, age, social background; culturally, they mostly reflect the values of developed western countries \cite{zhao2024gender, arzaghi2024understanding, ranjan2024comprehensive} in their decisions and generations, a problem yet to be fully tackled. Moreover, the most popular models are proprietary and their behavior is poorly documented, which could create a dangerous dependence on private companies if they were used to assist democratic processes \cite{feldstein2023consequences}. 
Democratic decision-making requires a fully intelligible and transparent process that humans can supervise, judge, and modify. Each step must be criticizable with a clear entity to hold accountable.

Still, LLMs bring new opportunities for monitoring democratic activities. In particular for participatory democracy and citizen consultations, such as Taiwan’s \textit{vTaiwan} platform\footnote{\url{info.vtaiwan.tw/}} (2014), the European Citizens’ Consultations\footnote{\url{oecd-opsi.org/innovations/the-european-citizens-consultations-eccs/}} (2018), or France's \textit{Grand Débat National}\footnote{\url{https://granddebat.fr/}} (Great National Debate, 2019), LLMs can help decipher the large amount of contributions through topic-modeling, summarization, and community detection \cite{small2023opportunities, galariotisArtificialIntelligenceThreatening2024, guembour-etal-2025-semantic}. 
Previous works have explored the usage of LLMs in democratic deliberations analysis \cite{tessler2024ai,fishGenerativeSocialChoice2025}. However, these studies focus on relatively small assemblies of up to a hundred participants, with clear on-topic statements. \citet{small2023opportunities} explore larger consultations, but rely on LLM-based\footnote{Using Anthropic's Claude model \citep{bai2022training}.} clustering and analyses. 

Large-scale consultations are notably difficult to process due to the diversity and sometimes noisiness of real-world citizen contributions  \cite{guembour-etal-2025-semantic}, which can simultaneously address multiple topics and policies, mixing-up facts, vindications, criticisms, proposals, and demands. The consequences are twofold: the synthesis or aggregation of a set of diverse contributions is difficult to explain or inspect; it requires extremely powerful NLP tools, capable of seamlessly handling such textual variability -- at the expense, however, of transparency and with increased dependence on very large, closed-weights LLMs. 


To avoid this problem, we propose the novel task of \textbf{Corpus Clarification}, illustrated in Figure~\ref{fig:explaination}. It aims at refining the initial corpus into a set of clear mono-topic opinions that are easier to process, analyze and inspect, even with small models. We first define the multiple steps of the task in Section~\ref{part:task}.
 We then manually annotate French contributions to the \textit{Grand Débat National} for this task. Our AI-human collaborative annotation process, detailed in Section~\ref{part:annotation}, ensures high-quality annotations. It also allows us to evaluate the capacities of multiple LLMs. We explore the resulting corpus \textbf{GDN-CC} in Section~\ref{part:corpus}. We then evaluate the capacities of locally-runnable models for Corpus Clarification in Section~\ref{part:experiments} and show that, when finetuned, they can perform on par or better than large API-based LLMs. We use these models to create \textbf{GDN-CC-large}\footnote{datasets and models available at \url{https://huggingface.co/collections/LequeuISIR/gdn-cc}} by automatically annotating 240k contributions of the \textit{Grand Débat National} as a resource for both political scientists and NLP researchers. Finally, we showcase the importance of \textit{Corpus Clarification} on an opinion clustering task, obtaining better clusters after clarification than with the original contributions.

%% file: parts/related_works.tex
\paragraph{LLMs and citizen consultations}
Recent works have explored LLMs for enhanced deliberation, from both experimental and ethical standpoints. \citet{small2023opportunities} conducted an extensive exploration of the usage of LLMs for deliberation as a complement to the Polis platform \cite{small2021polis},\footnote{\url{pol.is/}} focusing on topic modeling, summarization, prediction of votes and moderation. They highlighted the importance of human supervision in these tasks, for example by validating summaries to ensure accuracy and fairness. \citet{lazar2024can} argue that LLMs should be used to enhance \textit{non-instrumental} values of democracy, values that are intrinsically worthwhile such as deliberating or informing \citep{anderson2009democracy}.
However, as highlighted by \citet{revelAIFacilitatedCollectiveJudgements2025}, LLM-based systems for democratic deliberation exist, but they focus on small-scale processes.
\citet{tessler2024ai} identify common ground in groups of five citizens using an iterative human feedback loop grounded in Jürgen Habermas's theory of deliberative democracy \cite{habermas1985theory}. \citet{fishGenerativeSocialChoice2025} use LLMs to aggregate opinions for groups of up to 100 participants within a framework based on Social Choice theory \cite{sen1986social}. Although built on strong theoretical backbones, these works explore AI for deliberation in controlled and strongly constrained scenarios, which are difficult to scale to real-life large-scale deliberations. 

Existing work on democratic discourse either (i) focuses on small, manually curated datasets \cite{tessler2024ai, fishGenerativeSocialChoice2025}, or (ii) relies on end-to-end LLM-based aggregation and analysis \cite{small2023opportunities}. In contrast, our work focuses on scalable processing that enables traditional, more transparent NLP methods on large corpora, without involving LLMs in the decision loop. This work brings together political science and philosophical recommendations on the use of AI for democracy \cite{revelAIFacilitatedCollectiveJudgements2025,galariotisArtificialIntelligenceThreatening2024, lazar2024can} with the technological constraints of analyzing large-scale consultation corpora.

\paragraph{Argument mining}
Many recent works explore LLMs for opinion argument mining \cite{chen-etal-2024-exploring-potential,guida-etal-2025-llms}. This is for instance the case of \citet{ding-etal-2023-score, favero-etal-2025-leveraging} who applied it to the evaluation of opinion essays in education. Our work is also specifically related to argument mining in text written by citizens rather than argumentation professionals. Many works study this setting in tweets \cite{schaefer-stede-2020-annotation, dutta2020changing,iskender2021argument} or in citizen consultations \cite{liebeck-etal-2016-airport, fierro-etal-2017-200k,romberg-conrad-2021-citizen}. Additionally, The Key Point Analysis (KPA) task \cite{friedman-etal-2021-overview} aims at extracting the most prominent key-points representative of sets of opinions, either as an extractive or abstractive task. Finally, The concept of \emph{argumentative units} segmentation has been explored in previous work, e.g \cite{ajjour-etal-2017-unit} which used Bi-LSTMs to segment essays and editorials, and use a four-class span classification framework (claims, premises, anecdotes and assumptions).  We contribute to the argument mining literature with a manually-annotated and a large automatically-annotated corpora in French, using a three-class span classification designed specifically for citizen consultations and to be actionable for downstream tasks. 




\paragraph{Citizen Consultations Datasets}
In the last few years, multiple works have published citizen consultations: Polis \citep{small2021polis} openly shared twenty of their consultations.\footnote{\url{github.com/compdemocracy/openData/}} 
 Similarly, CoFE \cite{barriere-etal-2022-cofe} contains 4.2k proposals on the \textsl{Conference on the Future of Europe} consultation\footnote{\url{commission.europa.eu/strategy-and-policy/priorities-2019-2024/new-push-european-democracy/conference-future-europe_en}} and the Debating Europe dataset \cite{barriere-etal-2022-debating} contains 2.6k opinions from 18 debates related to the European Green Deal.\footnote{\url{commission.europa.eu/strategy-and-policy/priorities-2019-2024/european-green-deal_en}} \citet{romberg-conrad-2021-citizen} published five public consultations in German related to mobility transitions. More broadly, political tweets corpora on specific subjects have been released \cite{li-etal-2021-p,iskender2021argument,bilal-etal-2022-template,fourati-etal-2024-politun}, which are relatively close to citizen consultations. 

Our work brings a new large-scale opinion corpus, the first processed specifically for democratically-acceptable downstream tasks. We share both a high-quality manually annotated dataset of 1,231 French contributions, as well as an additional set of 240k automatically annotated contributions from the \textsl{Grand Débat National}, respectively \textbf{GDN-CC} and \textbf{GDN-CC-large}. The resulting corpus is, to our knowledge, the largest annotated democratic citizen consultation corpus shared to the community, and the only one in French.  



%% file: parts/corpusclarif.tex


The analysis of citizen consultations' corpora strongly relies on aggregating similar opinions together \cite{small2023opportunities,galariotisArtificialIntelligenceThreatening2024, guembour-etal-2025-semantic}. However, 
ethics-oriented works do not take into account the noisy reality of such free-form corpora \cite{galariotisArtificialIntelligenceThreatening2024}, and practical works rely heavily on proprietary LLMs to group \cite{small2023opportunities} or aggregate \cite{tessler2024ai, fishGenerativeSocialChoice2025} opinions. We designed the task of Corpus Clarification to turn a large noisy set of opinions into a clear standardized dataset that can be processed with more traditional and explainable NLP approaches such as embedding-based clustering. We achieve this transformation through a structured, three-step process, illustrated in Figure~\ref{fig:explaination}, and detailed below.

The first step is the \textbf{Argumentative Units (AU) extraction}. This step aims to identify the various themes or points covered within a contribution. This process breaks down the source material into smaller, distinct Argumentative Units, each focused on a single coherent topic. AUs are not necessarily contiguous, and some segments of the text might not be part of any AU.

The second step is the \textbf{Argumentative Structure (AS) detection} inside argumentative units. This task identifies the specific discourse types present within each Argumentative Unit, which provides fine-grained information for argument mining analysis.  
While the argument mining literature mostly uses a claim-premise classification \cite{chen-etal-2024-exploring-potential}, we adopt a three-type classification to better align with the requirements of downstream tasks that focus mostly on either policy proposals or citizens' feelings:

\begin{itemize}[itemsep=0pt, parsep=1pt, partopsep=0pt, topsep=0pt]
\item \textbf{Statement}: The expression of a sentiment or an observation;
\item \textbf{Solution}: A proposal for action or policy change;
\item \textbf{Premise}: A segment providing justification, evidence, or an example to support a statement or solution.
\end{itemize}

While \textit{Solutions} function similarly to \textit{claims} in traditional argument mining, we adopt this specific terminology to eliminate the ambiguity between \textit{statements} and \textit{claims} that often arises in citizens' contributions.

The final step, \textbf{Argumentative Unit Clarification}, aims at both forming context-independent argumentative units and normalizing the style to remove linguistic markers of the writers. Before clarification, AUs may lack self-sufficiency if they rely on context provided outside their boundaries (such as in the preceding text). The clarification process adds this necessary context to ensure that every AU is a fully comprehensible text, independent from its original contribution. Moreover, this step standardizes texts to a shared style and quality, similar to the task of style transfer \cite{toshevska2025llm}. This ensures that all citizens are handled similarly in downstream tasks. This is crucial since writing clarity was shown to be related to socioeconomic status \cite{dolek2018comparison}.

%% file: parts/annotation_process.tex
\subsection{The \textit{Grand Débat National} corpus}

The \textit{Grand Débat National} (GDN) 
is a nationwide citizen consultation held in France in 2019. The consultation prompted citizens to express their views across four main themes: \textit{Taxation and public spending}, \textit{Organization of the state and public services}, \textit{Democracy and citizenship}, and \textit{Ecological transition}. A significant portion of this consultation involved online questionnaires, each concluding with a critical open-ended prompt: "Do you have anything to add about [theme]?". Our starting point are the responses to these open-ended questions. These specific contributions are valuable as they allow citizens to freely express their opinions. The complete anonymized corpus is publicly available for download, use and sharing via the French \textit{Grand Débat} platform.\footnote{https://granddebat.fr/pages/donnees-ouvertes}

\paragraph{Data Preparation} The original dataset comprises 355k individual contributions. To ensure data quality and relevance, we applied several filtering steps. After removing duplicate entries, we excluded contributions shorter than 30 characters or longer than 600 characters to focus on responses with meaningful content that were not excessively long for annotation, resulting in 240k contributions. 
Finally, aiming for diversity in response complexity, we ensured a uniform distribution across the four themes and across lengths of contributions. We used the French sentencizer provided by the \texttt{spaCy} library \cite{vasiliev2020natural} to estimate the number of sentences in each text, as sentences were used as an approximation of opinions in a previous analysis \cite{guembour-etal-2025-semantic}.

\subsection{Annotation guidelines and process}
Five  native French speakers specializing in political science were hired to annotate 300 contributions each. The annotation was performed in two main steps: segmentation and clarification.

\paragraph{Contributions segmentations} The tasks of \textit{Argumentative unit segmentation} and \textit{Argumentative structure detection} were performed jointly. Annotators were instructed to parse the raw text and identify three primary segment types: Premises, Statements, and Solutions. These segments were grouped into topically coherent Argumentative Units. This combined segmentation and detection process was designed to accelerate the annotation process and to ensure that each AU contains only salient, argument-relevant information. 

\paragraph{Argumentative units clarification} Once the initial segmentation step was complete, the argumentative units were processed for clarification by a randomly selected LLM in a pool of four. 
The resulting clarifications were then displayed to annotators, who were asked to revise them if necessary. 
This hybrid human-LLM process allowed us to speed-up the annotation. It also yielded useful data to compare the ability of the various LLMs systems to clarify argumentative units.     

Annotation was performed in two phases. In the first phase, annotators could regenerate clarifications using different LLMs to maximize annotation quality. During the second phase, which took place during the annotation of the last quarter of the contributions, 
annotators were asked to modify the first generated clarification, even when its quality was poor. This allowed us to characterize the types and frequency of errors made by LLM systems, which we report in Section~\ref{subsec:types_errors}.
\newline
The process totaled 1,553 annotations of 1,231 individual contributions, the remaining allowing for the computation of inter-annotator agreement. We describe the annotators' training and monitoring in Appendix \ref{subapp:annotators}.


\subsection{Annotation tool}

As there was, to our knowledge, no existing tool suitable for our two-step annotation process, we implemented our own.
This interface\footnote{available at \url{https://github.com/LequeuISIR/GDNAnnotationPlatform}} enables the annotators to perform segmentation and rewriting in a single step, so that the clarification can benefit from the understanding of the contribution gained during segmentation. To generate the initial automatic clarifications, we relied on four families and sizes of models: \texttt{GPT-4.1} \cite{achiam2023gpt}, \texttt{Qwen-3-32b} \cite{yang2025qwen3}, \texttt{Llama-3.1-8b} and \texttt{Llama-3.3-70b}  \cite{grattafiori2024llama}. 
More information regarding the annotation tool, its implementation and the annotation process is in Appendix~\ref{subapp:platform}.

%% file: parts/dataset.tex
\subsection{Statistics of the dataset}
GDN-CC consists of 1,231 contributions that have been manually-annotated for the Corpus Clarification task. Table~\ref{tab:stats_segmentation} reports basic statistics regarding this dataset. Overall, GDN-CC contains significantly more solutions (57.9\%) than statements (21.2\%) and premises (20.9\%). This result is expected, as citizens who are actively going to the consultation platform expect to be heard and to influence the decision-making process. Interestingly, the \textit{Democracy and Citizenship} theme contains fewer argumentative units, and in proportion fewer solutions (54\% versus 58-60\%) and more statements (24\% versus 19-21\%) than the other themes. Participants may struggle to express solutions for these less-discussed topics, resulting in more descriptive statements. The number of argumentative units per contribution spans from one to eleven, and the number of argumentative segments from one to twelve. This highlights the complex and variable nature of the corpus.


\begin{table*}[ht]
  \centering
  \begin{tabular}{lc|c|ccc}
    \hline
    \textbf{Theme} & \textbf{contribs.} & \textbf{AUs}  & \textbf{statements} & \textbf{solutions} & \textbf{premises} \\
    \hline
    \textbf{Taxation and Public Spending} & 312 & 594 & 175 \tiny{(18.7\%)} & 556\tiny{(59.6\%)} & 202 \tiny{(21.7\%)} \\
    \textbf{Ecological Transition} & 305 & 590 & 187 \tiny{(20.5\%)} & 543 \tiny{(59.5\%)} & 183 \tiny{(20.0\%)} \\
    \textbf{Organization of the State} & 308 & 577 & 197 \tiny{(21.4\%)} & 532 \tiny{(57.9\%)}& 190 \tiny{(20.7\%)}\\
    \textbf{Democracy and Citizenship} & 306 & 524 & 212 \tiny{(24.2\%)} & 474 \tiny{(54.2\%)} & 187 \tiny{(21.4\%)}\\
    \hline
    \textbf{Total}  & \textbf{1231} & \textbf{2285} & \textbf{771} \tiny{(21.2\%)}& \textbf{2105} \tiny{(57.9\%)}& \textbf{762} \tiny{(20.9\%)}\\
    \hline
  \end{tabular}
  \caption{\label{citation-guide}
    Number of contributions, argumentative units, statements, solutions, and premises per theme and in total in \textbf{GDN-CC}, the manually-annotated corpus. Percentages are calculated row-wise.
  }
   \label{tab:stats_segmentation}
\end{table*}

\subsection{Discrepancies and agreements between annotators \label{ssec:agreements}}

We evaluated the inter-annotator agreement separately for the two  tasks of argumentative unit (AU) extraction and argumentative structure (AS) detection.

\paragraph{AU extraction} For AU extraction, we rely on a span-limitation based metric, WindowDiff \cite{pevzner-hearst-2002-critique}, used in previous studies for inter-annotator agreements of span annotations \cite{javorsky-etal-2025-mockconf}. It uses a sliding window of size $k$ ($k = 15$ tokens in our case) and evaluates, at each step, how many boundaries the two annotations have within the window. 
These are aggregated in a probability of boundary disagreement within a sliding window, with lower values indicating greater consistency of the segmentation. Following recent works on segmentation \cite{ding-etal-2023-score, favero-etal-2025-leveraging}, we also compute a token-overlap metric: Given two spans $S_1$ and $S_2$ annotated by two annotators, we evaluate $s(S_1, S_2) = \min(\frac{|S_1|\cap|S_2|}{|S_1|}, \frac{|S_1|\cap|S_2|}{|S_2|})$ for each pair of spans in the annotation. To ensure optimal matching, we solve the span-pair assignment problem using the \texttt{SciPy} package\footnote{using the \texttt{linear\_sum\_assignment} method} \cite{virtanen2020scipy}. A match is defined if $s(S_1, S_2)$ is above a set threshold $\lambda$.

We observe a strong level of agreement between annotators with both metrics. The mean and median WindowDiff values over all annotations are 0.09 and 0.08 respectively. Similarly, the token overlap metric yields micro- and macro-F1 of 0.72 and 0.77 for a 50\% overlap threshold ($\lambda = 0.5$), and 0.42 and 0.44 for perfect alignment ($\lambda = 1$).
 Table~\ref{tab:agree_segmentation} in Appendix~\ref{subapp:interannot} also reports precision and recall. 


\paragraph{AS detection} 
For AS detection, we compute the inter-annotator agreement independently of the extracted argumentative units. We considered the task as a token tagging problem: for each token, annotators classified it as \textit{statement}, \textit{premise}, \textit{solution}, or did not classify it (considered a \textit{none} class). We compute the \emph{Cohen's kappa} of the classification task and obtain $\kappa = 0.54$, corresponding to a moderate agreement according to  \citet{landis-etal-1977-measurement}. We also compute the F1 score for the three classes, and find thar the agreement varies significantly between classes, reflecting the nature of citizen speech. We found the strongest agreement for \textit{solutions}, reaching $F1=0.80$. 
 This category is the most critical for downstream policy analysis, and its high reliability demonstrates that annotators can consistently identify actionable citizen proposals. However, scores for \textit{premises} and \textit{statements} were lower (0.61 and 0.52 respectively). This discrepancy stems from a structural confusion between the two classes: in non-expert discourse, a sentiment or observation (\textit{statement}) frequently acts as a justification (\textit{premise}) for a solution. Such confusions were similar to those observed in previous argument mining studies \cite{ding-etal-2023-score}. By merging statements and premises into a single class, we obtain a 0.78 score for the joint class. Although we kept the three-class granularity in the rest of this study, future work could consider merging them into a single class. Examples of confusions between statements and premises, and the confusion matrix for the three classes is in Appendix \ref{subapp:interannot}. 



\subsection{Evaluating the capacities of models for Clarification}
During the first annotation phase, we considered four models: \texttt{GPT-4.1}, \texttt{Qwen-32b}, \texttt{Llama-70b}, and \texttt{Llama-8b} for AU clarification. Annotators could iterate through multiple AI-generated clarifications from randomly-chosen LLMs before accepting the one they edit. Observing this process allows us to derive a comparative evaluation of these models using a probabilistic model of annotators' behavior. In summary, we treat the unknown quality of each model $l$ as a parametric distribution $Q_l$. 
A user accepts the LLM clarification if $Q_l$ exceeds a threshold $\tau_k$, which varies with the number of clarification attempts $k$.
If an output is accepted, its quality $q_l$ is observed via the ROUGE-L score against the final human-validated text. 
By jointly optimizing the parameters of $Q_l$ and $\tau_k$ to maximize the likelihood of user actions, we found that \texttt{GPT-4.1} performed best ($\mu=0.94$), closely followed by \texttt{Llama-70B} ($\mu=0.93$) and \texttt{Qwen-32B} ($\mu=0.92$) which proved to be close competitors for zero-shot clarification. \texttt{Llama-8B} ($\mu=0.90$) yielded good performances despite its limited size. This finding provides a first hint at the abilities of smaller models for the task, that we further observe in Section~\ref{part:experiments}. A detailed explanation of the probabilistic model and its estimation is in Appendix~\ref{app:opti}.

\subsection{Qualitative exploration of the clarifications corrections}
\label{subsec:types_errors}

Although the first phase of the annotation delivered high-quality clarifications and an evaluation of the LLMs systems, it made it difficult to perform a qualitative analysis of clarification errors. This is because annotators were permitted to reject some low-quality LLM outputs in their revision process. 
The second annotation phase required annotators to accept (and edit when necessary) the first automatically generated clarification, irrespective of its quality. We manually examined 100~argumentative units annotated during this phase (25 for each LLM) for which the clarification was edited by the annotator. We identified four main types of errors:
\begin{itemize}[itemsep=0pt, parsep=1pt]
    \item \textbf{Over-analysis}: LLM added analysis or a conclusion absent from the original text;
    \item \textbf{Miscomprehension}: LLM did not fully regenerate the opinion, by being wrong or by omitting important information;
    \item \textbf{Over-specificity}: LLM added information present in the text but removed by the annotator, such as unimportant details or content from other AUs;
    \item \textbf{Misformulation}: LLM output contained the right information, but its phrasing was unnatural or contained an extraneous introductory phrase.\footnote{Such as "here is the argument: **\{\{argument\}\}**".}
\end{itemize}

Table~\ref{tab:errors_examples} in Appendix~\ref{app:corpus_analysis} presents an example of each type of error, while Table \ref{tab:results_errors} reports the proportion of each error for each model. Over-specificity, miscomprehension and misformulation are seldom found (respectively 3\%, 12\% and 13\% of all errors) but all four models consistently struggle with over-analysis (72\% of all errors), 
 even when specifically prompted not to add any information not expressed in the text. This behavior is highly detrimental in the context of democratic processes, especially when models add justifications that were not present in the author's contribution. Complementing this finding, further analyses presented in Appendix~\ref{subapp:caraterization_corrections} suggest that a significant part of the human annotation effort of clarifications was to remove information added by the LLMs.

%% file: parts/experiments.tex
Focusing on having locally-runnable systems and not relying on large proprietary LLMs, we explore the use of different systems to automate the tasks of AU extraction and AS detection. In particular, we evaluate the capacities of Small Language Models (SLMs, $\leq10B$ parameters), including their instruct and task-specific finetuned variants. For all three tasks, we thus evaluated \texttt{Llama-3.1-8B} \cite{grattafiori2024llama}, \texttt{Mistral-7B-v0.3} \cite{Jiang2023Mistral7}, \texttt{Qwen2.5-7B} \cite{team2024qwen2} and \texttt{Gemma-2-9b} \cite{team2024gemma} families of models. 
We used 70\% of the data for training and 15\% each for validation and testing. We used \texttt{GPT-4.1} as a comparison baseline for all tasks. 
Results are reported in Table~\ref{tab:experiments_results}. For finetuned models, we report the results of the best-performing learning rate. See Appendix~\ref{app:expe} for finetuning details and figures.\footnote{all experiments are available at \url{https://github.com/LequeuISIR/GDNCorpusClarification}}

\begin{table*}[ht]
\centering
\footnotesize 
\setlength{\tabcolsep}{2pt} 
\begin{tabularx}{\textwidth}{l l c | YYY | YYY | YYY | YYY | c}
\toprule
\multirow{2}{*}{\textbf{Task}} & \multirow{2}{*}{\textbf{Metric}} & \multicolumn{1}{c}{\textbf{GPT-4.1}} & \multicolumn{3}{c}{\textbf{Llama-8b}} & \multicolumn{3}{c}{\textbf{Mistral-7b}} & \multicolumn{3}{c}{\textbf{Qwen-7b}} & \multicolumn{3}{c}{\textbf{Gemma-9b}} & \multicolumn{1}{c}{\textbf{Encoder}} \\
\cmidrule(lr){3-3} \cmidrule(lr){4-6} \cmidrule(lr){7-9} \cmidrule(lr){10-12} \cmidrule(lr){13-15}  \cmidrule(lr){16-16}
& & - & Base & Inst. & Ft. & Base & Inst. & Ft. & Base & Inst. & Ft. & Base & Inst. & Ft. & - \\
\midrule

\multirow{2}{*}{AU Extract.} & Micro-F1 & \underline{0.75} & 0.09 & 0.54 & 0.61 & 0.04 & 0.62 & 0.71 & 0.46 & 0.66 & \underline{0.75} & 0.03 & 0.67 & 0.66 & \textbf{0.82}\\
                         & Macro-F1 & 0.73 & 0.10 & 0.59 & 0.72 & 0.03 & 0.58 & 0.78 & 0.40 & 0.66 & \underline{0.81} & 0.04 & 0.69 & 0.76 & \textbf{0.83}\\
\addlinespace

\multirow{2}{*}{AS Detect.} & Micro-F1 & \textbf{0.76} & 0.12 & 0.31 & 0.65 & 0.02 & 0.17 & 0.70 & 0.34 & 0.56 & 0.67 & 0.00 & 0.49 & \underline{0.73} & 0.70 \\
                        & Macro-F1 & 0.76 & 0.12 & 0.36 & 0.71 & 0.01 & 0.15 & \underline{0.78} & 0.37 & 0.64 & 0.75 & 0.00 & 0.57 & \textbf{0.79} & 0.76 \\
\addlinespace

\multirow{2}{*}{AU Clarif.} & BERTScore & 0.81 & 0.64 & 0.79 & \underline{0.85} & 0.26 & 0.77 & \textbf{0.86} & 0.65 & 0.82 & \underline{0.85} & 0.59 & 0.80 & \textbf{0.86} & -\\
                        & Rouge-L & 0.45 & 0.19 & 0.39 & \underline{0.56} & 0.05 & 0.33 & \underline{0.56} & 0.18 & 0.52 & 0.54 & 0.21 & 0.39 & \textbf{0.60} & - \\

\bottomrule
\end{tabularx}
\caption{Performance metrics across models and tasks with  the best values for each metric in \textbf{bold} and the second best values \underline{underlined}. Low performance of Base models are due to their inability to consistently generate the expected output format.}
\label{tab:experiments_results}
\end{table*}

\subsection{Argumentative Units segmentation}

Given a written contribution to the citizen consultation, the system has to extract the spans of texts corresponding to different argumentative units. More specifically, we prompt the models to output argumentative units as a list. Table~\ref{tab:experiments_results}'s first row displays the Macro-F1 and Micro-F1 using the token overlap metric used for inter-annotator agreement (Section~\ref{ssec:agreements}). 
While the base models mostly failed this task, instruct models perform better, with supervised finetuning further improving the results. Interestingly, encoder-based models outperformed decoder models including \texttt{GPT-4.1} for this task when train as a BIO (\textit{beginning, Inside, Outside} a span, further explained in appendix~\ref{subsubapp:encoder_AU}) token tagger. \texttt{Cambertav2-base} \cite{antoun2024camembert20smarterfrench} displayed Micro-F1 and Macro-F1 of 0.82 and 0.83.

\subsection{Argumentative Structure detection}
Given an argumentative unit, the system has to extract and classify the spans as \textit{statement}, a \textit{premise} or a \textit{solution}. We provide the LLMs both the initial contribution and the argumentative unit from which spans need to be extracted.
The second row in Table~\ref{tab:experiments_results} shows the Macro-F1 and Micro-F1 using a label-constrained version of the token overlap metric used for AU segmentation:
$$
s_{cons}(S1,S2) = \delta_{label(S1),label(S2)}\times s(S1,S2),
$$
where $\delta_{i,j}=1$ if $i=j$ else $0$. As for argumentative units segmentation, non-instruct SLMs are unable to perform this task, while instruct models achieve much better results. Regarding finetuned models, \texttt{Gemma-2-9B} (Macro-F1: 0.79, Micro-F1: 0.73) outperforms other SLMs and performs on par with \texttt{gpt-4.1} (Macro-F1: 0.76, Micro-F1: 0.76). Experiments with encoder-based models as a span extraction and tagging task (detailed in Appendix~\ref{subsubapp:encoder_AU}) showed that while giving lower results, \texttt{Cambertav2-base} proved effective for resource-constrained scenarios (Macro-F1: 0.76, Micro-F1: 0.70).


\subsection{Argumentative Unit clarification}
Given an argumentative unit, the system has to rewrite it as a clear, self-sufficient argument by extracting only the important information from the text. We provided the models with the initial contribution for context and the argumentative unit to clarify. 
We report BERTScore\footnote{Using \texttt{bert-base-multilingual-cased} as backbone.} \cite{zhang2019bertscore} and Rouge-L\footnote{using the \texttt{rouge-score} 0.1.2 package} \cite{lin-2004-rouge} for all models in Table~\ref{tab:experiments_results}'s last row.
While all finetuned SLMs achieve comparable performance in terms of BERTScore, slightly exceeding \texttt{GPT-4.1}, more pronounced disparities are observed for the ROUGE-L metric. We hypothesize that informational content remains consistent across models but those with higher ROUGE-L scores produce a semantic structure that more closely aligns with the style of human annotators, explaining lower scores of non-finetuned models. We find \texttt{Gemma-2-9B} (BERTScore: 0.86, ROUGE-L: 0.60) to outperform other systems including \texttt{GPT-4.1}.

Using \texttt{GPT-4.1-nano} as a judge, we compared finetuned clarifications against initial LLM outputs. Finetuned \texttt{Gemma-2-9b} was preferred 66\% of the time, initial outputs 26\% of the time. Additionally, we manually overviewed 100 clarifications from \texttt{Gemma-2-9b} for the four types of errors introduced in Section~\ref{subsec:types_errors}. We find that finetuning significantly mitiged the \emph{over-analysis} issue, from 19\% in \texttt{GPT-4.1} outputs down to 2\%. The miscomprehension error is the most common (12\%), but mostly stems from over-simplifying the argumentative unit rather than adding false information. Over-specificity and misformulation (3 examples each) are seldom found. The last three types are less concerning than the over-analysis for democratic processes, as they do not add unexpressed information. See Appendix~\ref{app:llmasjudge_clarifs} for the evaluation prompt and comparison examples between \texttt{GPT-4.1} and \texttt{Gemma-2-9b} for which over-analysis was corrected.

\subsection{Overview of SLMs clarification errors}

%% file: parts/downstream.tex
\subsection{Implication for downstream tasks}
Clustering similar opinions is a key step in many downstream tasks such as topic modeling or summarization \cite{small2023opportunities}.

To assess the potential improvements for democratic downstream tasks, we evaluate the effect on clustering quality of (i) AU segmentation and (ii) AU clarification. We use UMAP \cite{mcinnes-etal-2018-umap} for dimension reduction and HDBSCAN 
\cite{mcinnes-etal-2017-hdbscan} for clustering on the three types of texts -- namely, the initial contributions, the extracted AUs, and the associated clarifications.   
We evaluate clustering quality using (1) unsupervised clustering metrics, namely the silhouette \cite{rousseeuw-etal-1987-silhouettes} and Davies-Bouldin (DB) \cite{davies-etal-2009-cluster} scores, and (2) \verb|GPT-4.1-nano| as a judge in a pairwise comparison setup. \newline
For the former, we compare the clusters created from the argumentative units and from their clarifications on each of the four themes of the consultation. The results showed a consistent improvement in both scores when using the clarifications instead of their associated argumentative units: the average silhouette score went from 0.46 to 0.59 (higher is better) and DB score from 0.48 to 0.46 (lower is better). 
\newline
For the latter, we sample pairs of texts coming from the \emph{same} cluster. The model is then presented with pairs from two different clusterings (contributions, AUs, or clarification); the task is to identify the most coherent pair based on thematic specificity and consistency. We tested three settings with 100~sampled pairs per theme:
\begin{enumerate}[itemsep=0pt, parsep=1pt, partopsep=0pt, topsep=0pt]
\item Initial contributions vs.\ Argumentative units;
\item Argumentative units vs.\ Clarifications;
\item Argumentative units vs.\ Argumentative units, using the clusters based on clarifications.
\end{enumerate}

The third setting serves as a control to ensure the judge’s preference is driven by thematic coherence rather than by the improved syntax of clarifications.

The LLM judge showed a near-total preference for the more granular units in all settings. In Setting~1, argumentative units were prefered over raw contributions in 84\% of cases. Setting~2 showed a 91\% preference for clarifications. Setting~3 confirmed this trend (90\% preference), showing that the coherence gain persists even when evaluating the same text type (AU) rearranged into clusters from clarifications. All results are statistically significant ($p < 0.001$ in all settings). Prompts and statistical analyses are given in Appendix~\ref{app:clustering}.

%% file: parts/full_corpus.tex
\subsection{Annotating GDN-CC-large}
Having shown that finetuned SLMs were reliable for the task of Corpus Clarification, and that this task was actually helping clustering and potentially other downstream tasks, we applied our pipeline using the best-performing finetuned SLMs (\texttt{Qwen-7b} for AU extraction and \texttt{Gemma-9b} for AS detection and AU clarification)\footnote{Further experiments with finetuned encoders displayed better results for AU extraction, and another version of GDN-CC-large will be shared.} to the full corpus consisting of 240k contributions, yiedling a dataset of 300k argumentative units, split into 155k statements, 282k solutions and 62k premises. Table \ref{tab:stats_final_corpus} displays the corpus statistics per theme and in total. This annotation reinforce the analysis made in Section \ref{part:corpus}, showing that \textit{Democracy and Citizenship} contains proportionally fewer solutions and more statements. We can also see that \textit{Ecological Transition} and \textit{Taxation and Public Spending} got significantly more contributions (72.4\% of all contributions  and 73.6\% of all argumentative units in the corpus) than the two other themes. These themes are strongly related to the context in which the \textit{Grand Débat National} was organized, following the \textit{Yellow Vest} movement which started due to a raise in taxes.


%% file: parts/Appendix.tex
\section{Annotation process and platform}
\label{app:annotation}
\subsection{Annotation Platform}
\label{subapp:platform}
The annotation platform was developed in javascript, using the React.js\footnote{\url{react.dev/}} and Next.js frameworks\footnote{\url{nextjs.org/}}. The back-end server to handle contributions distributions, annotator accounts, and saving the annotations was developed in Python, using the Flask framework \cite{grinberg2018flask}. For the automatic clarification, we use the Open AI's API\footnote{\url{platform.openai.com/}} for \texttt{GPT-4.1} and Groq inference provider\footnote{\url{groq.com/}} for the three other models.

\begin{figure*}
    \centering
    \includegraphics[width=0.99\linewidth]{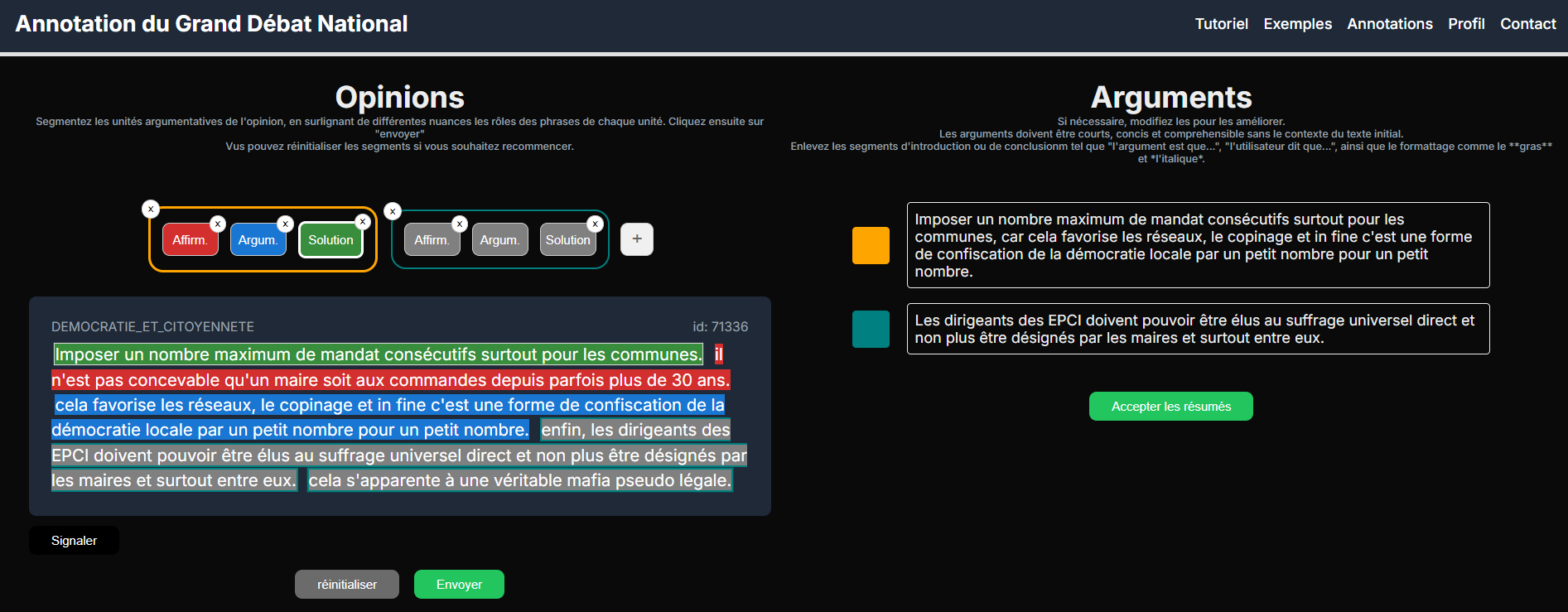}
    \caption{Annotation interface. On the left is the segmented contribution, and on the right the clarification for each argumentative unit.}
    \label{fig:interface_annotation}
\end{figure*}

\begin{figure*}
    \centering
    \includegraphics[width=0.99\linewidth]{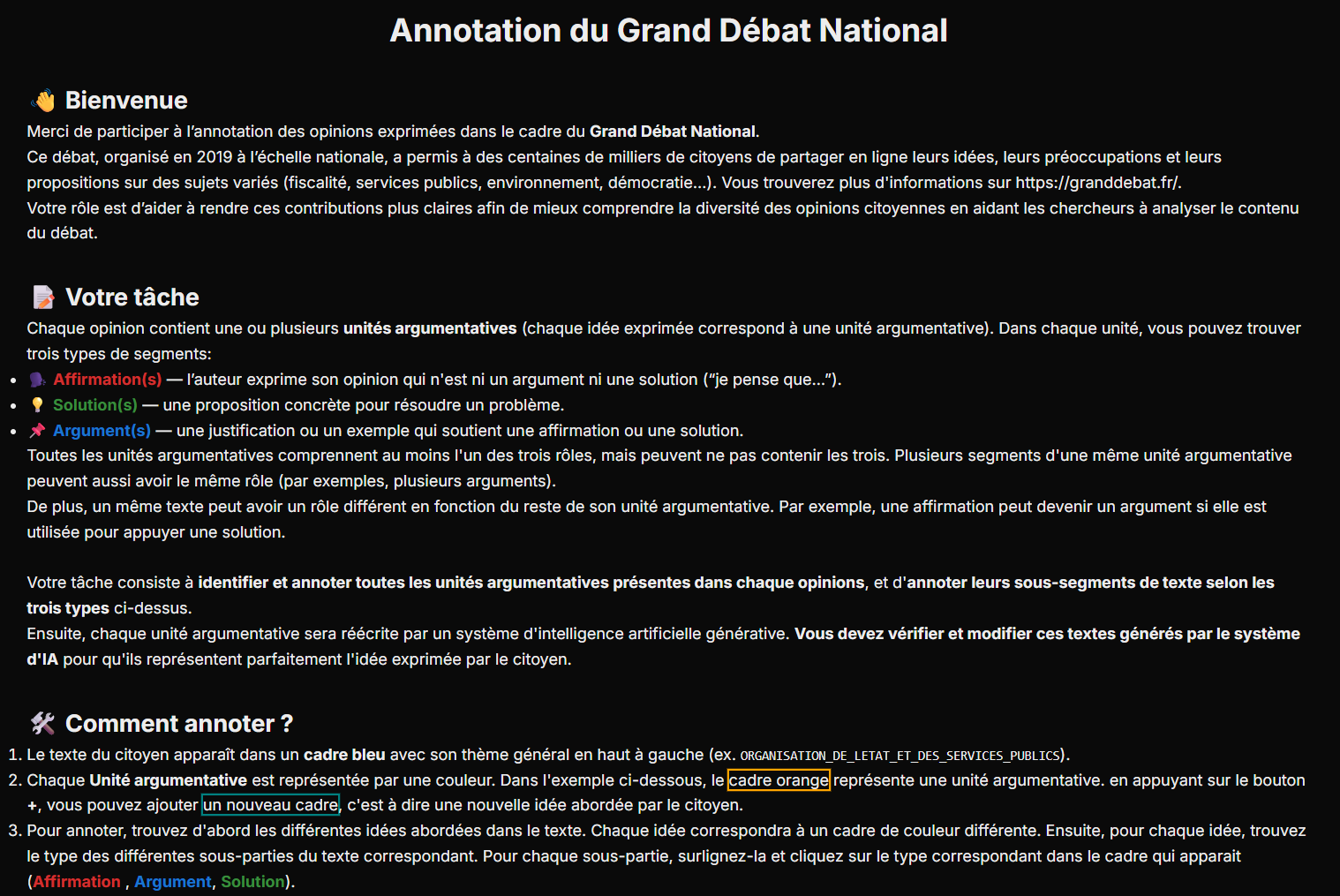}
    \caption{Welcome page. See Figure~\ref{fig:explanation_translation} for translation of the task explanation.}
    \label{fig:interface_welcome}
\end{figure*}

\begin{figure*}
    \centering
    \includegraphics[width=0.99\linewidth]{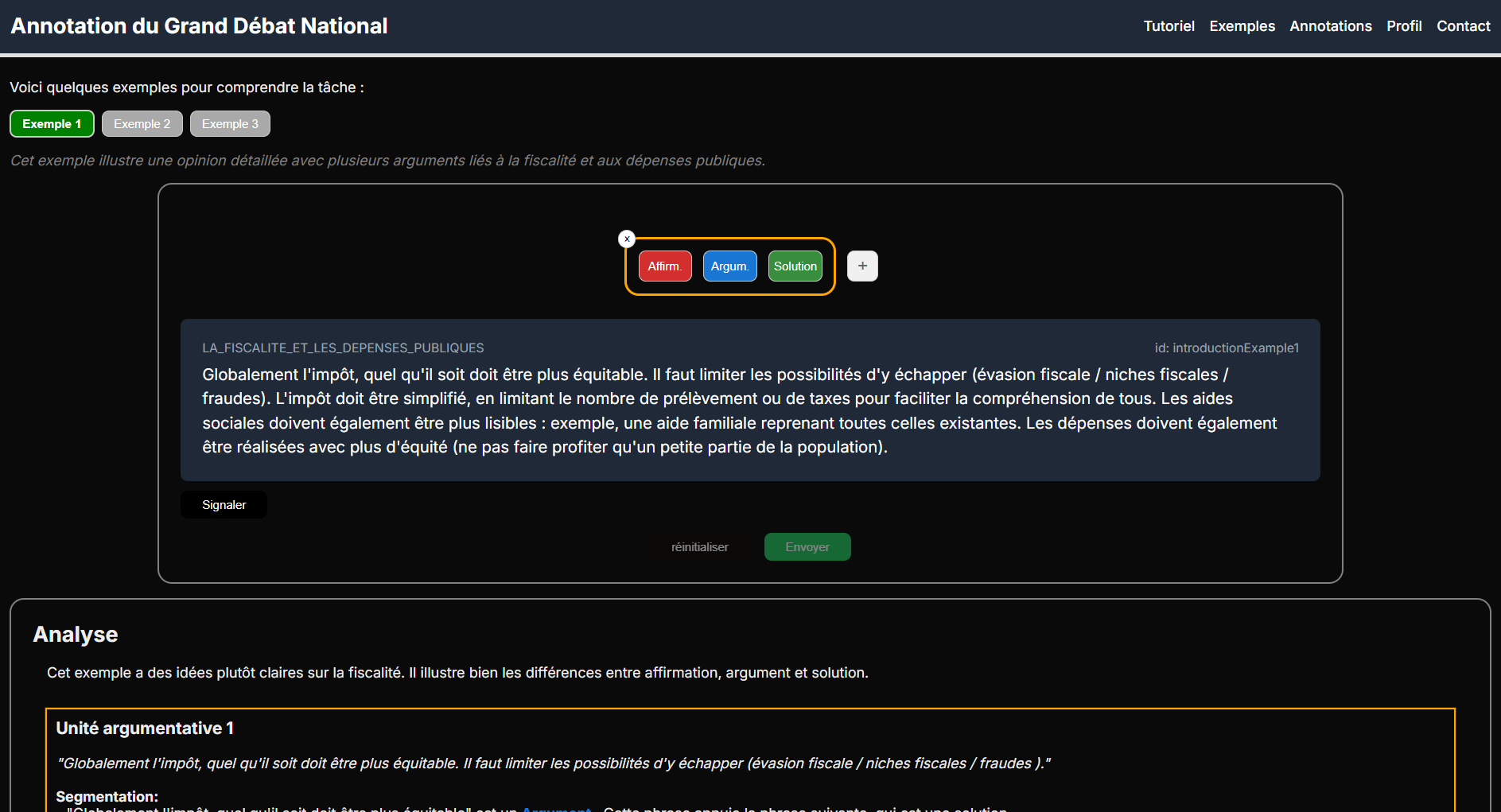}
    \caption{Example page.}
    \label{fig:interface_example}
\end{figure*}

\begin{figure*}
    \centering
    \includegraphics[width=0.99\linewidth]{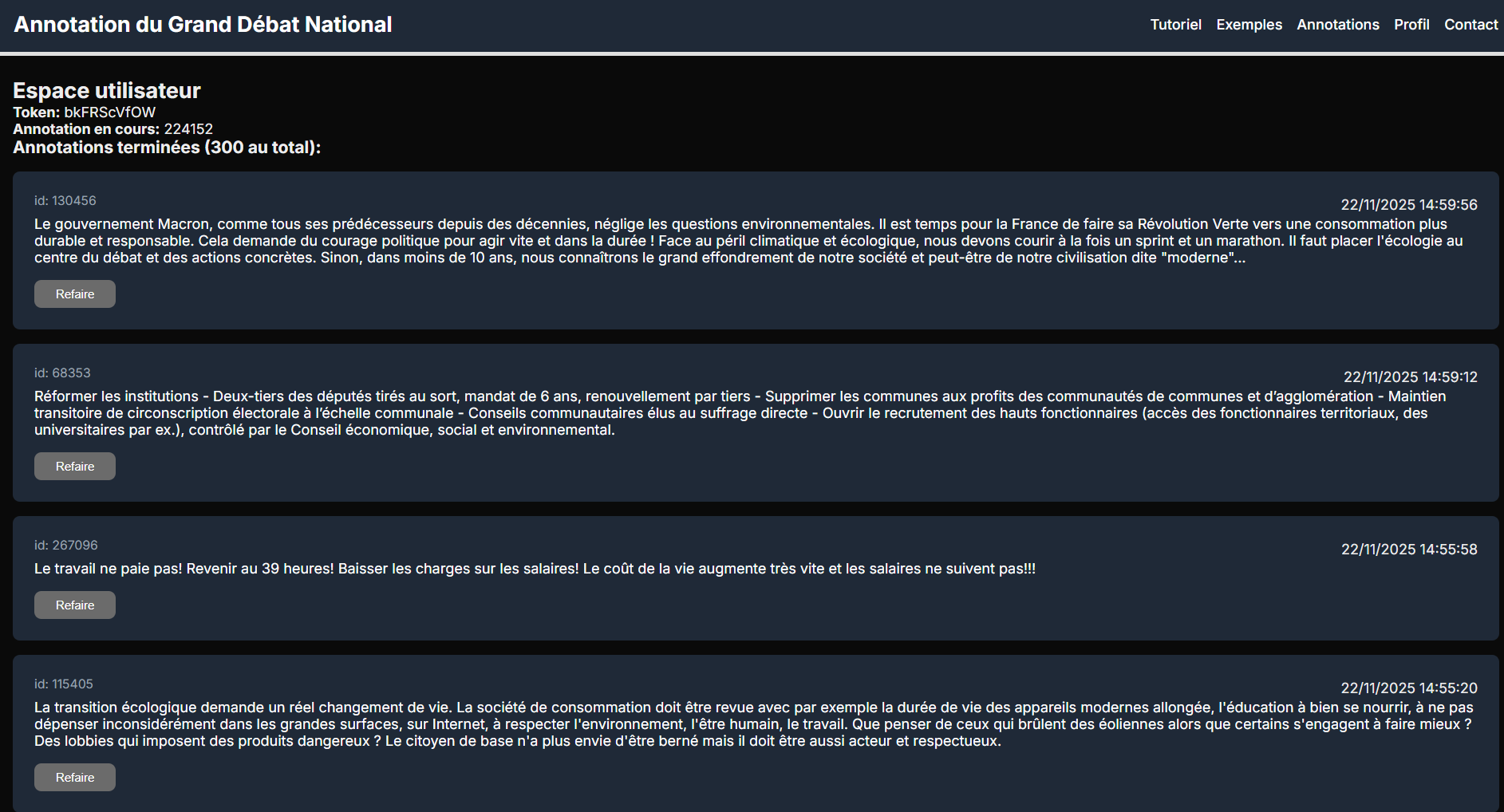}
    \caption{Account page.}
    \label{fig:interface_account}
\end{figure*}

\begin{figure*}
    \centering
    \includegraphics[width=0.99\linewidth]{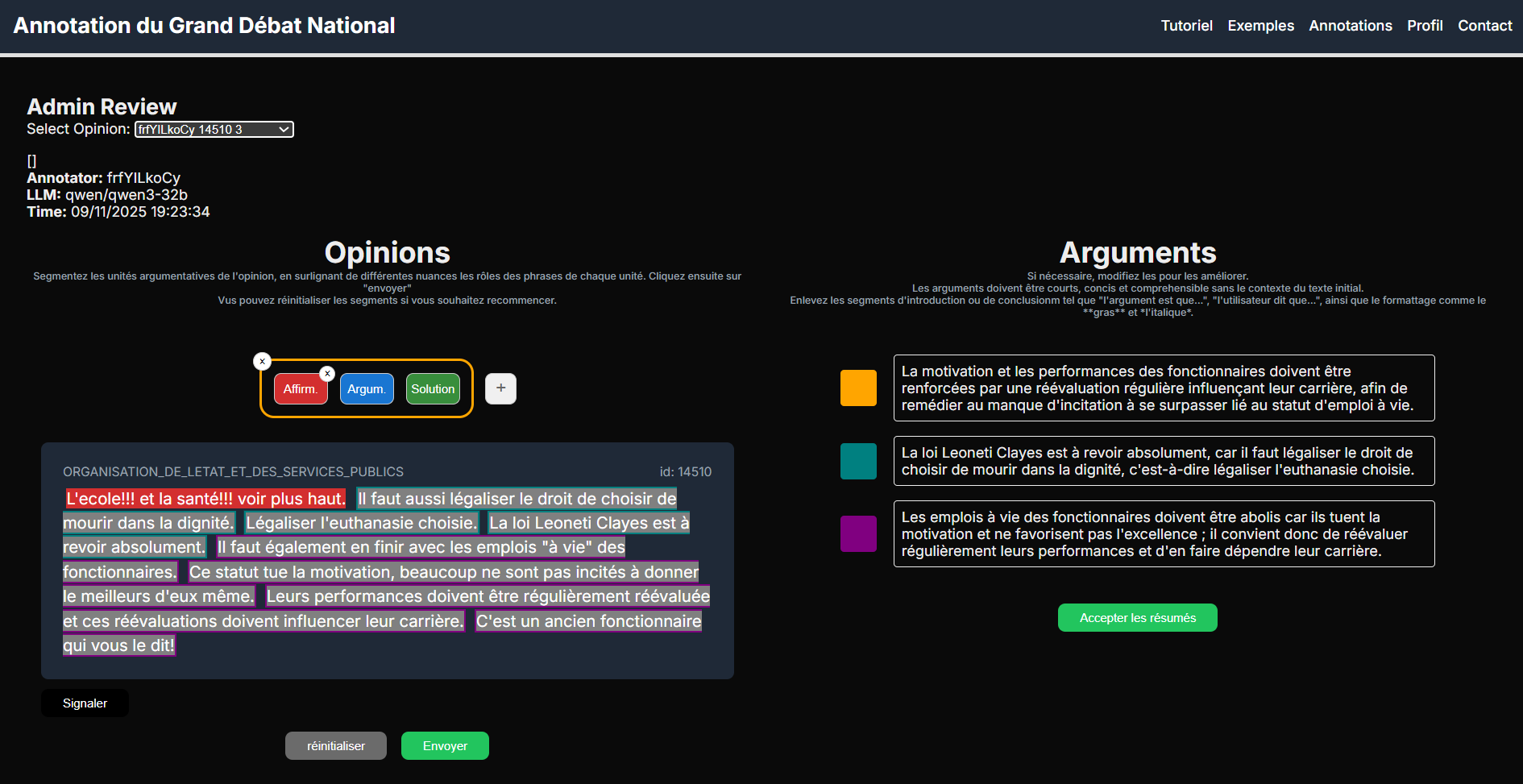}
    \caption{Administration page.}
    \label{fig:interface_admin}
\end{figure*}

The platform contains a welcome/tutorial page accessible to all (Figure \ref{fig:interface_welcome}), as well as an example page to see some annotations already done (Figure \ref{fig:interface_example}). Annotators are given a unique token. They must validate three examples for which the expected annotation is given and explained before having access to the annotation interface shown in Figure \ref{fig:interface_annotation}. Annotators also have access to an account page (Figure \ref{fig:interface_account}), from which they can see the number of annotations they have done so far, as well as all the examples they annotated. They are allowed to go back to previous annotations to re-do them if needed. The interface also implements an administration page for the allowed tokens to see all done annotations (Figure \ref{fig:interface_admin}). During the annotation, annotators were allowed to report and skip contributions to annotate if they were not understandable, contained hate speech, were too long, or if they contained personal information.

We share our code for the platform to the community\footnote{In the camera-ready version of the paper}. While all written parts are in French, They can easily be modified to fit other annotations needs in other languages. 

\begin{figure*}[ht]
  \begin{promptbox}
  \begin{verbatim}
Each opinion contains one or more argumentative units (each idea expressed 
corresponds to one argumentative unit). In each unit, you can find 
three types of segments:

- Statement(s): the author expresses their opinion, which is neither an argument nor 
a solution (“I think that...”).
- Solution(s): a concrete proposal to solve a problem.
- Argument(s): a justification or example that supports a statement or solution.

All argumentative units include at least one of the three roles, but may not contain 
all three. Several segments of the same argumentative unit may also have the 
same role (for example, several arguments). In addition, the same text may have a 
different role depending on the rest of its argumentative unit. 
For example, a statement can become an argument if it is used to support a solution.

Your task is to identify and annotate all argumentative units present in each 
opinion, and to annotate their text sub-segments according to the three types 
above. Each argumentative unit will then be rewritten by a generative 
artificial intelligence system. You must check and modify these texts generated by 
the AI system so that they perfectly represent the idea expressed by the citizen.
\end{verbatim}
  \end{promptbox}
  \caption{Translation of the task explanation on the annotation platform.}
  \label{fig:explanation_translation}
\end{figure*}

\subsection{Annotators Training and Monitoring}
\label{subapp:annotators}
Annotators are graduate students in political science who were paid 25 euros an hour for the annotation (300 euros each in total), above the minimum wage in France. All annotators participated in a live presentation of the task before having access to the platform. The platform, tasks, and expectations were explained to them. Examples of gold annotations were provided to further explain the task. After the meeting, annotators had to annotate three controlled examples before starting the real annotation process. 

 After a third of the annotation was done, a meeting was held with four of the five annotators. The fifth annotator could not attend, but was sent a report of the meeting afterwards. The objective of this meeting was to discuss annotation difficulties and methods and to discuss some contributions for which two annotations by two different annotators were significantly different. This meeting greatly helped with improving the quality of annotations: The mean WindowDiff agreement went from 0.10 before the meeting to 0.08 (median 0.09 to 0.06). The token-overlap metric went from 0.74 and 0.69 macro-F1 and micro-F1 to 0.83 and 0.80 respectively. The mean and median accuracy of span types (\textit{statement}/\textit{premise}/\textit{solution}) went from 0.64 and 0.63 to 0.66 and 0.77. 

\subsection{Inter-Annotator Agreement}
\label{subapp:interannot}
\begin{figure}
    \centering
    \includegraphics[width=0.80\linewidth]{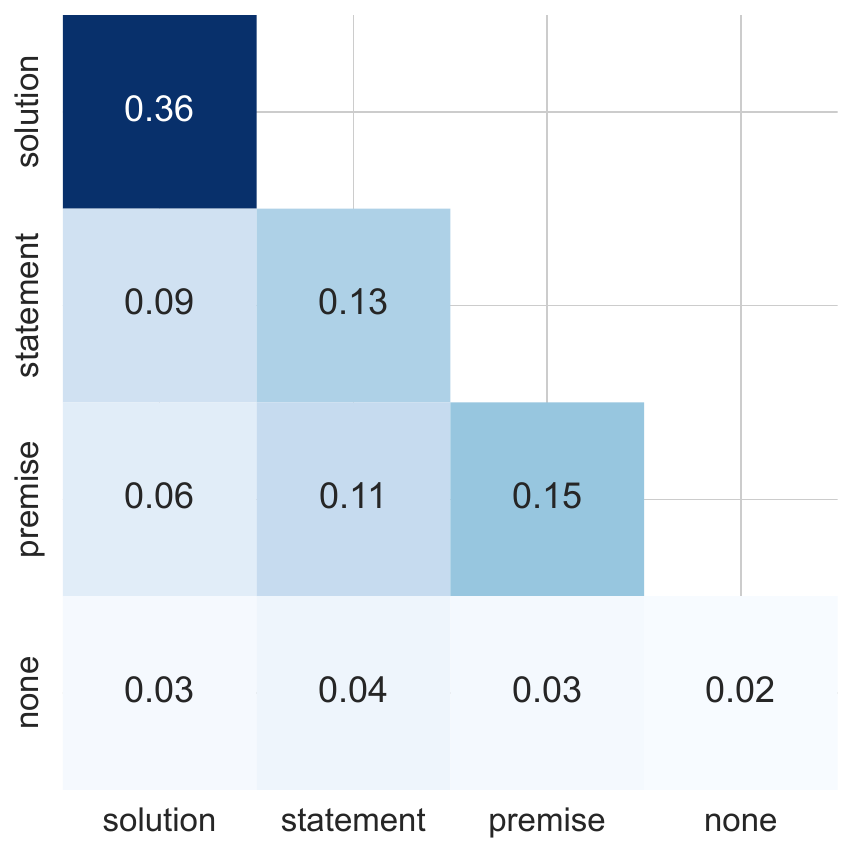}
    \caption{Proportions of tokens labels pairs in all contributions annotated by two different annotators.}
    \label{fig:matrice_confusion}
\end{figure}

Table \ref{tab:agree_segmentation} provides micro and macro precision, recall, and F1 for the task of Argumentative Units segmentation. Figure \ref{fig:matrice_confusion} shows the proportions of each annotated type pairs for tokens' argumentative type. We find that 36\% of the text tokens are annotated twice as \textit{solutions}, showing both the importance of this type and the strong agreement between annotators. This figure also displays the rather strong confusion between \textit{statements} and \textit{premises}. 

\begin{table}[H]
  \centering
  \begin{tabular}{ll|cc}
    \hline
    & & $\lambda=0.5$ & $\lambda=1$ \\
    \hline
    \textbf{precision} & micro & 0.78 & 0.45\\
     & macro & 0.80 & 0.42\\
     \hline
    \textbf{recall} & micro & 0.65 & 0.38   \\
    & macro & 0.76 & 0.41\\
    \hline
    \textbf{F1} & micro & 0.71 & 0.41\\
    & macro & 0.76 & 0.42\\
    \hline
  \end{tabular}
  \caption{
    Argumentative Units match between annotators.
  }
   \label{tab:agree_segmentation}
\end{table}

We provide some examples of confusions between \textit{statements} and \textit{premises}. All texts are originally French and translated to English. In many cases, the confusion is made when the citizen uses its own belief or experience as justification: 
\begin{itemize}[itemsep=0pt, parsep=1pt, partopsep=0pt, topsep=0pt]
    \item "\textit{Every country has its own history and customs.}" in  "\textit{Every country has its own history and customs. When integrating into a country, you must be able to respect and accept those of the host country.}";
    \item "\textit{Over 3 km near my house: 70-90-70-90-80-70-50}" in "\textit{You have to stop changing speed limits too often. Over 3 km near my house: 70-90-70-90-80-70-50}";
    \item \textit{"There have been more significant climate variations throughout history."} in "\textit{Let's stop focusing on humanity's ability to influence “climate change.” There have been more significant climate variations throughout history.}".
\end{itemize}

\section{Complementary Analysis of the Manually-Annotated Corpus}
\label{app:corpus_analysis}
\subsection{Statistics of the Corpus}
\label{subapp:manual_corpus_analysis}
\begin{figure*}
    \centering
    \includegraphics[width=\linewidth]{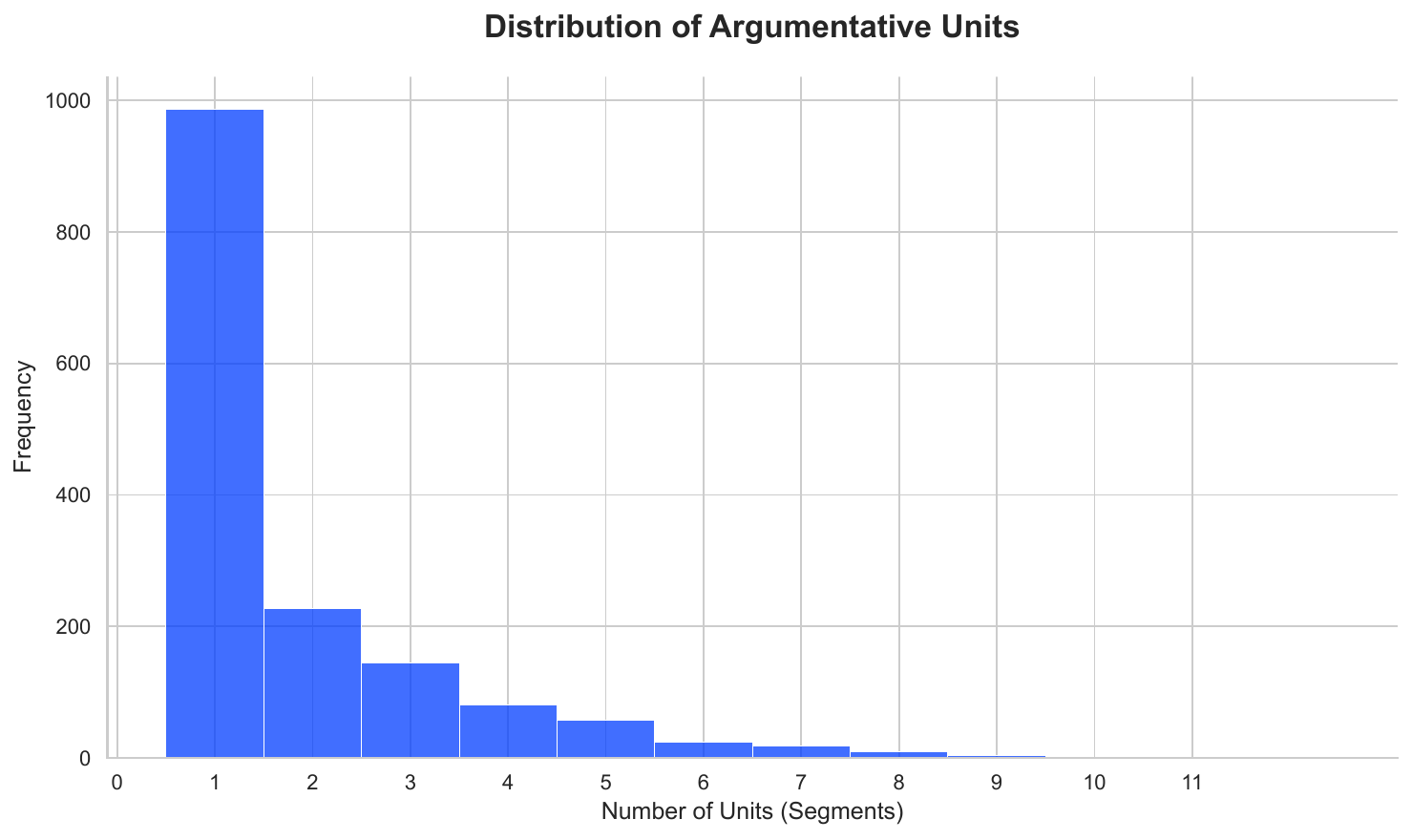}
    \caption{Distribution of numbers of Argumentative Units in the manually-annotated corpus.}
    \label{fig:distribution_AU}
\end{figure*}

\begin{figure*}
    \centering
    \includegraphics[width=\linewidth]{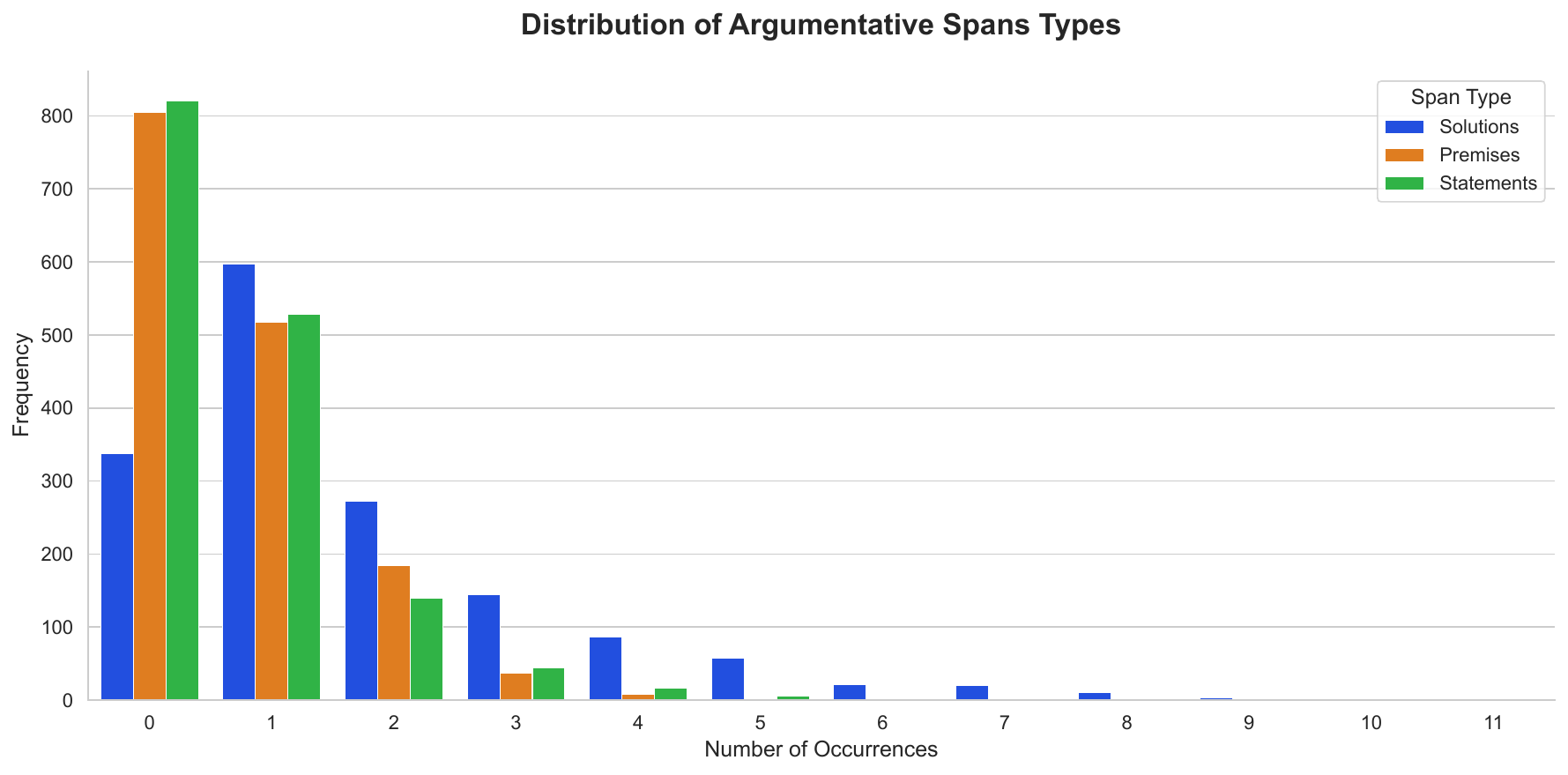}
    \caption{Distribution of numbers of argumentative spans types in the manually-annotated corpus.}
    \label{fig:distribution_spans}
\end{figure*}

In this section, we provide more statistics about the manually-annotated corpus. Figures \ref{fig:distribution_AU} and \ref{fig:distribution_spans} respectively displays the distributions of number of argumentative units and of argumentative spans types per contribution.

\input{parts/summary_corrections}

\subsection{Clarification Errors}
Table \ref{tab:errors_examples} provides an example of LLM output and annotator correction for each error type, in French and with their English translations. The information removed by the annotator is displayed in red while the information added by the annotator is shown in green. Table \ref{tab:results_errors} displays the proportion of types of errors for each model and in total found in the manually overviewed corrections.

\begin{table*}
  \centering
  \begin{tabular}{l|cccc||c}
    \hline
      & \textbf{GPT-4.1}  & \textbf{Qwen-32b} & \textbf{Llama-70b} & \textbf{Llama-8b} & \textbf{Total} \\
    \hline
    \textbf{\% of corrected output}  & 20.3\% & 27.8\% & 29.9\% & 29.7\% & 26.7\%\\
    \hline
    \hline
    \textbf{Over-analysis} & 76\% & 72\% & 80\% & 60\% & 72\%\\
    \textbf{Miscomprehension} & 8\% & 16\% & 0\% & 24\% & 12\%\\
    \textbf{Over-specificity} & 4\% & 4\% & 4\% & 0\% & 3\%\\
    \textbf{Misformulation} & 12\% & 8\% & 16\% & 16\% & 13\%\\
    \hline
  \end{tabular}
  \caption{
    Proportions of types of error per theme for each AI system during the second annotation phase.
  }
   \label{tab:results_errors}
\end{table*}

\begin{table*}
  \centering
  \begin{tabularx}{\textwidth}{ l|X|X }
    \hline
    \textbf{Error type}  & \textbf{Output} & \textbf{Correction} \\
    \hline
    \textbf{Over-analysis} & Réduire les frais de fonctionnement, les avantages et le nombre d’élus \textcolor{red}{permettrait de faire des économies significatives dans les dépenses publiques.}
    \newline
    \newline
    \small \textit{Reducing operating costs, benefits, and the number of elected officials \textcolor{red}{would lead to significant savings in public spending.}}
    & Il faut réduire les frais de fonctionnement, les avantages et le nombre d’élus.
    \newline
    \newline
    \small \textit{Operating costs, benefits, and the number of elected officials must be reduced.}
    \\
    \hline
    \textbf{Miscomprehension} & Développer un navigateur, un moteur de recherche et une messagerie open source pour concurrencer les géants de l'internet actuels.
    \newline
    \newline
    \small \textit{Develop an open source browser, search engine, and email service to compete with the current internet giants.}
    & Développer un navigateur, un moteur de recherche et une messagerie open source pour concurrencer les géants de l'internet actuels, \textcolor{green}{en mobilisant toutes les universités européennes sur ce projet.} 
    \newline
    \newline
    \small \textit{Develop an open source browser, search engine, and email service to compete with the current internet giants, \textcolor{green}{by mobilizing all European universities on this project.}}
    \\
    \hline
    \textbf{Over-specificity} & La solution est de respecter le vote des citoyens et le référendum, sans plan B, 
    \textcolor{red}{plutôt que de supprimer une ville entière pour des zadistes.} 
    \newline
    \newline
    \small \textit{The solution is to respect the citizens' vote and the referendum, without a plan B,  \textcolor{red}{rather than removing an entire city for the sake of protesters.}}
    & La solution est de respecter le vote des citoyens et le référendum, sans plan B.
    \newline
    \newline
    \small \textit{The solution is to respect the citizens' vote and the referendum, without a plan B. } \\
    \hline
    \textbf{Misformulation} & \textcolor{red}{L'argument clair et auto-suffisant sous-jacent est : **}Les plus pollueurs devraient payer plus, selon un système proportionnel à leur degré de pollution.\textcolor{red}{**} 
    \newline
    \newline
    \small \textit{\textcolor{red}{The clear and self-sufficient underlying argument is: **}The biggest polluters should pay more, according to a system proportional to their degree of pollution.\textcolor{red}{**}}
    & Les plus pollueurs devraient payer plus, selon un système proportionnel à leur degré de pollution.
    \newline
    \newline
    \small \textit{The biggest polluters should pay more, according to a system proportional to their degree of pollution.}\\
    \hline
  \end{tabularx}
  \caption{
    Example of AI system output and correction by the annotators for each type of error. In over-analysis, the removed text was not present in the initial contribution. In Over-specificity, it was present in the contribution but not in the argumentative unit.   
  }
  \label{tab:errors_examples}
\end{table*}

\section{Experiments}
In this section, we give details and complementary results of the experiments described in Section \ref{part:experiments}. 
\label{app:expe}
\subsection{Finetuning details}
\begin{figure*}
    \centering
    \begin{subfigure}{0.49\textwidth}
        \centering
        \includegraphics[width=\linewidth]{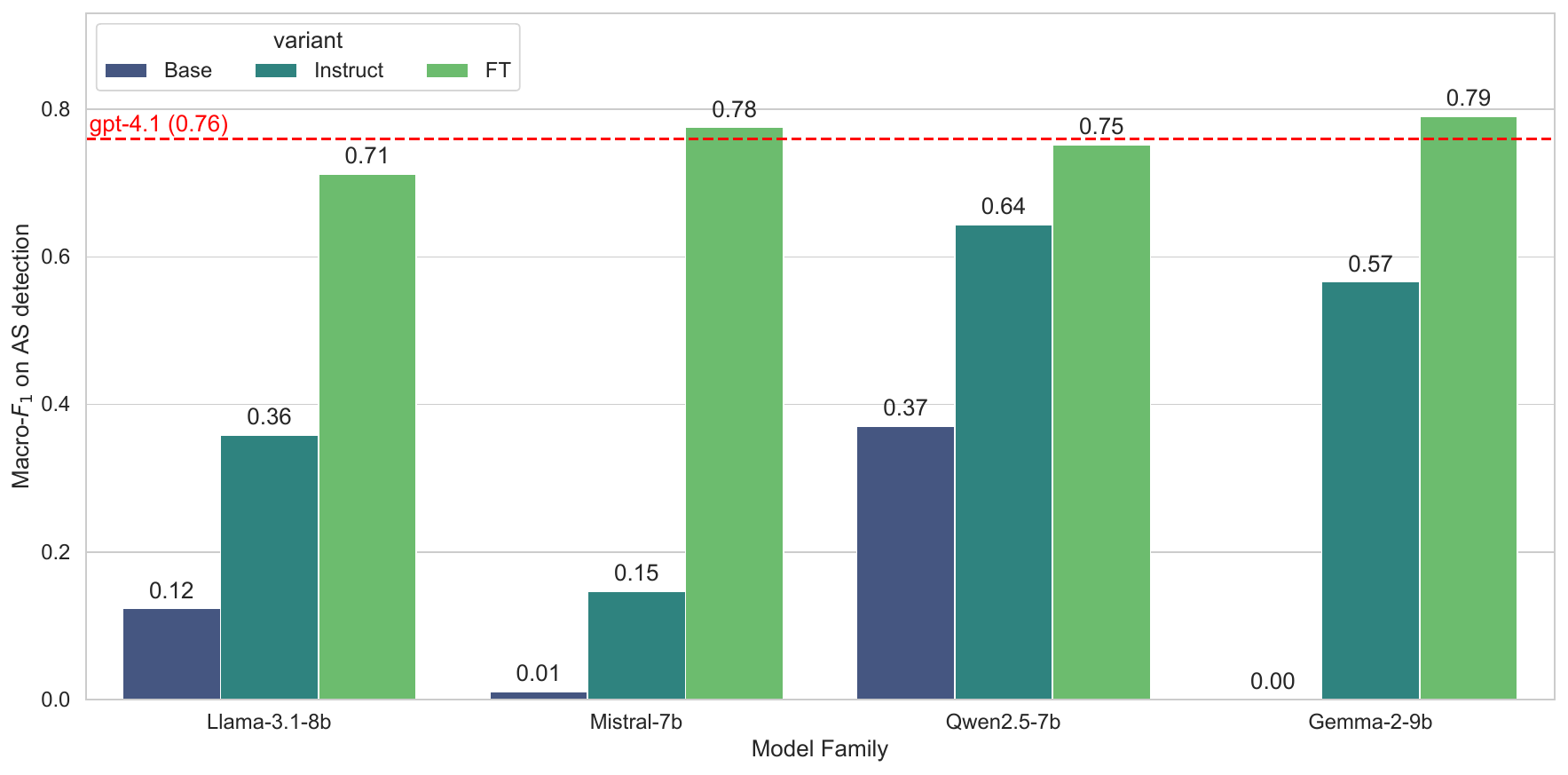}
        \caption{Macro-F1}
    \end{subfigure}
    \begin{subfigure}{0.49\textwidth}
        \centering
        \includegraphics[width=\linewidth]{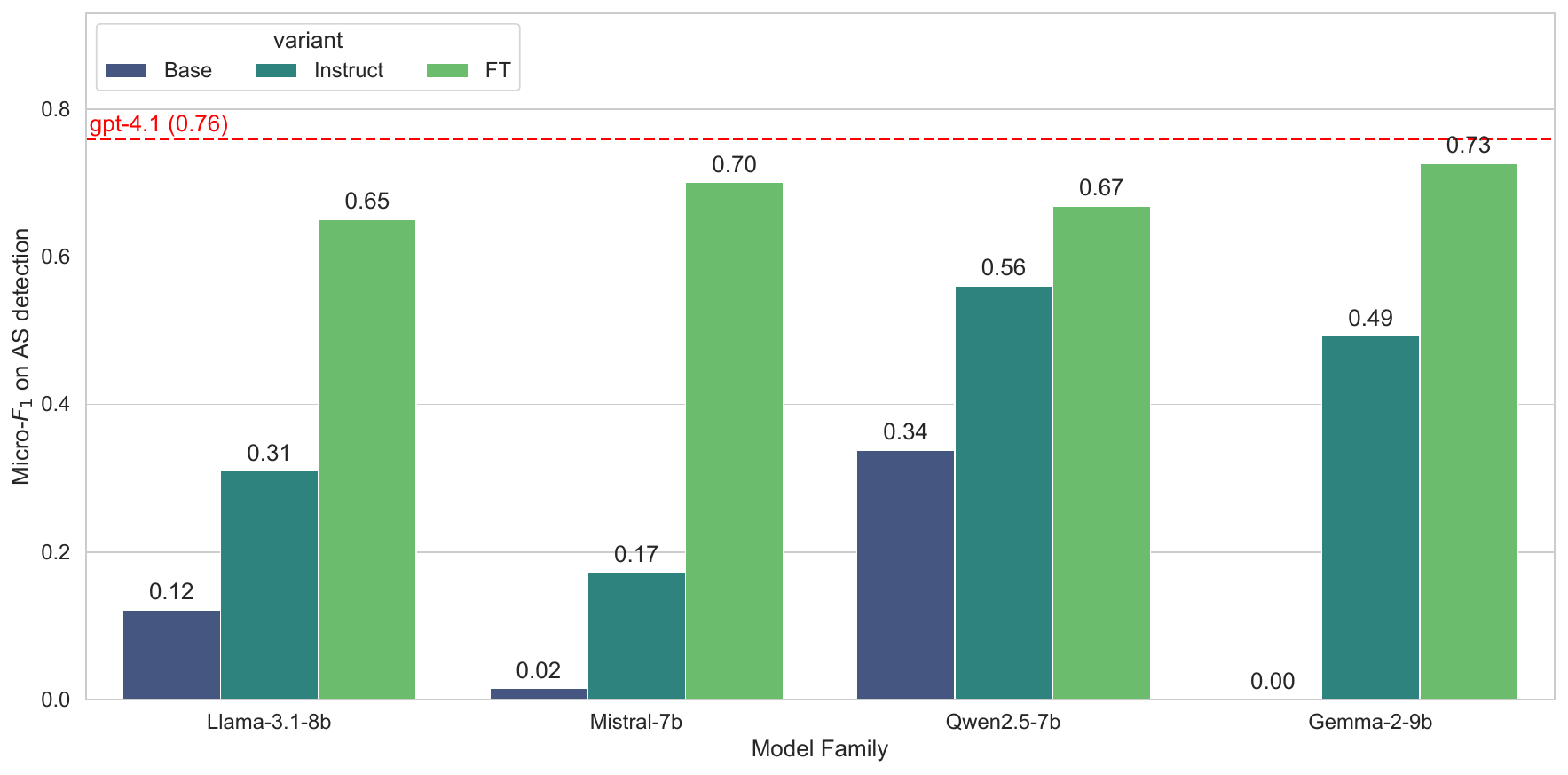}
        \caption{Micro-F1}
    \end{subfigure}
    \caption{Macro and Micro F1 of different model families on the task of Argumentative Structure detection}
    \label{fig:F1_as_extraction}
\end{figure*}

\begin{figure*}
    \centering
    \begin{subfigure}{0.49\textwidth}
        \centering
        \includegraphics[width=\linewidth]{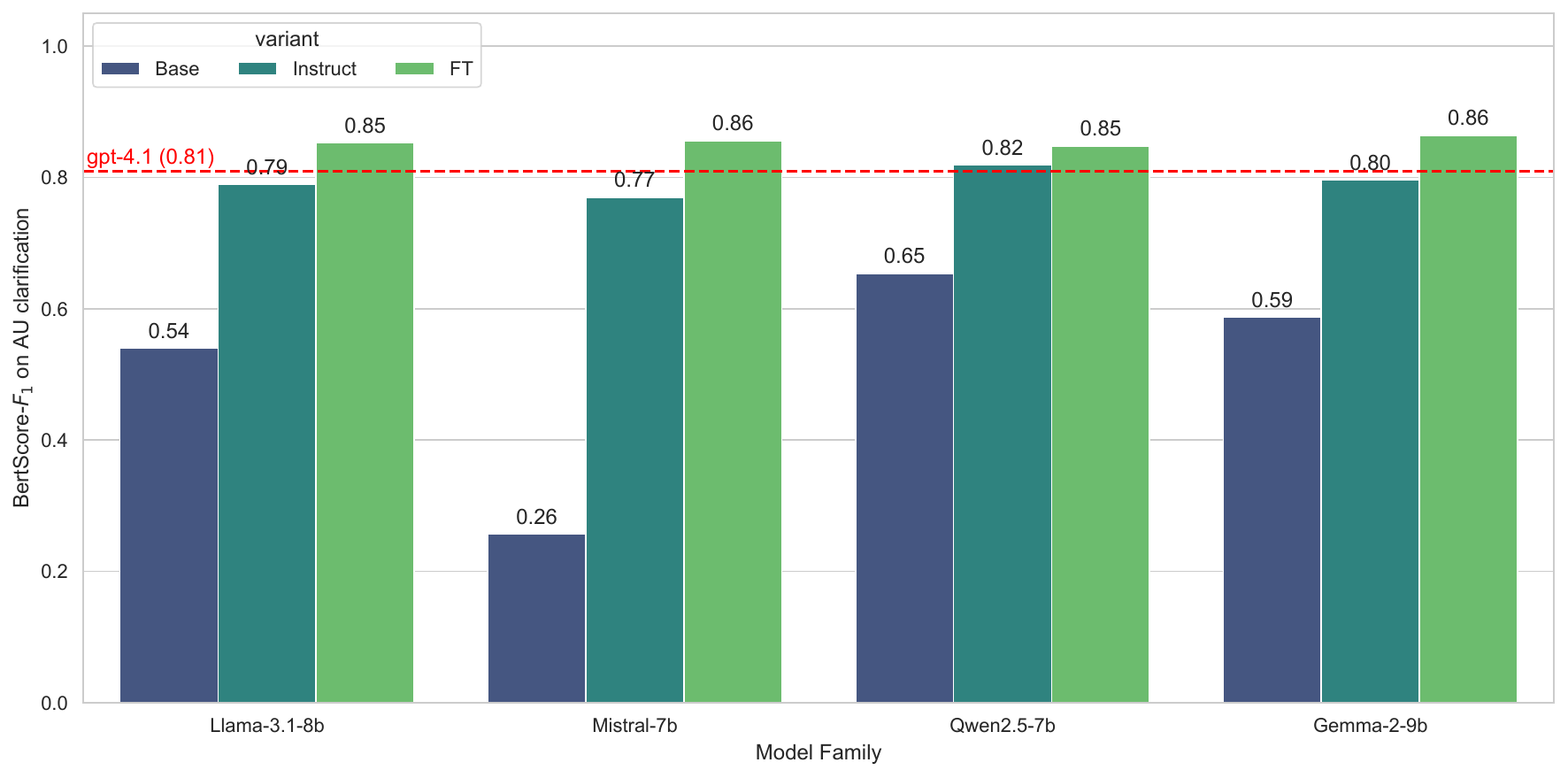}
        \caption{BERTScore}
    \end{subfigure}
    \begin{subfigure}{0.49\textwidth}
        \centering
        \includegraphics[width=\linewidth]{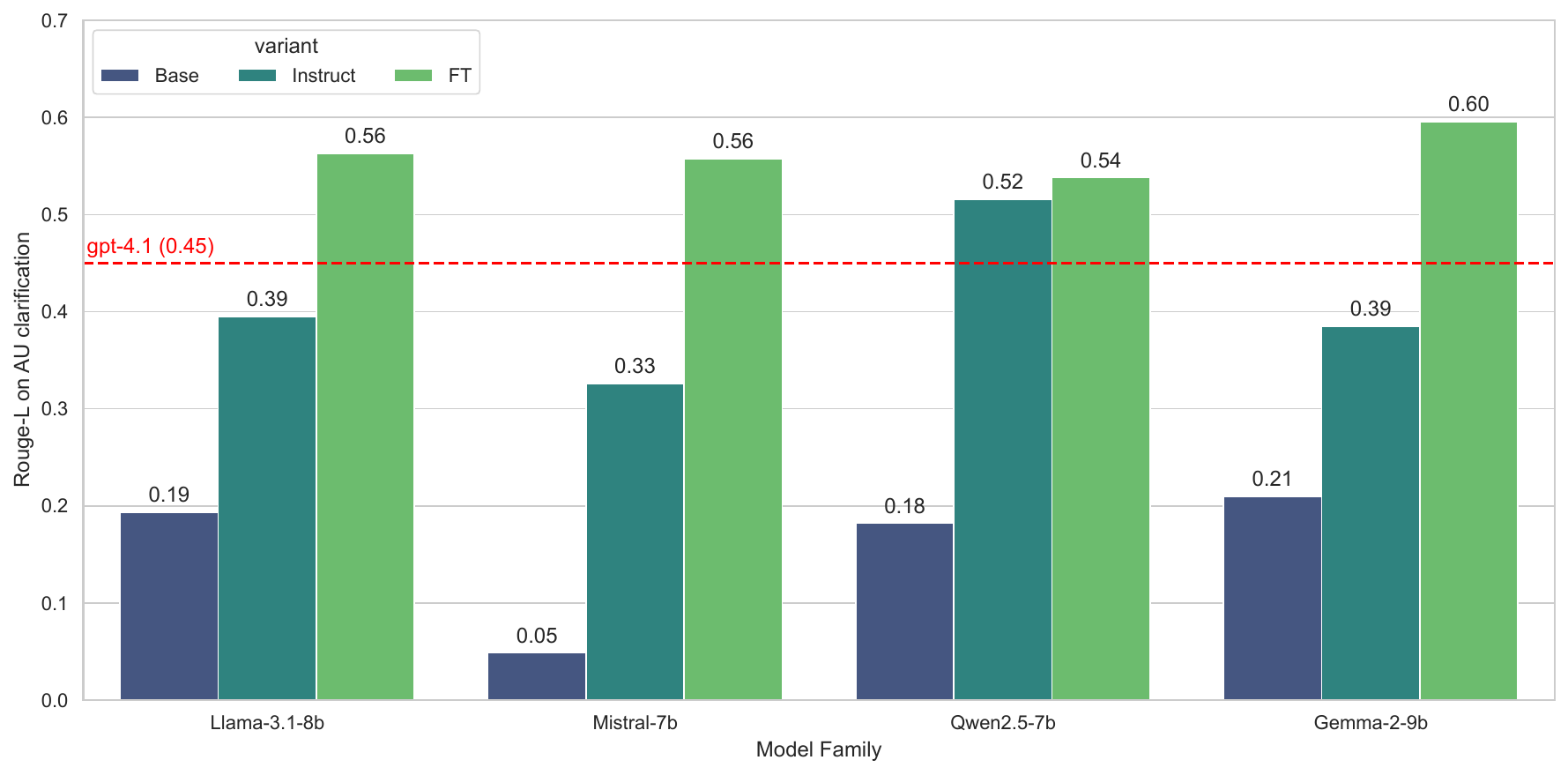}
        \caption{ROUGE-L}
    \end{subfigure}
    \caption{BERTScore and ROUGE-L of different model families on the task of Argumentative Unit clarification}
    \label{fig:bertscore_au_clarif}
\end{figure*}

All models are trained for up to 5 epochs with a batch size of $2$ and 8 steps of gradient accumulation. We report the best results for all model trying learning rates in $[1e-5, 3e-5, 5e-5, 1e-4]$. We set the learning rate warm-up ratio at 5\%. All models have an early stopping callback based on the evaluation set loss, and most stopped training after 2 to 3 epochs for all tasks. We use the TRL package \cite{vonwerra2022trl} for the supervised finetuning. All models were finetuned on one H100 GPU and evaluated using one A100 GPU, and the full finetuning and evaluation pipeline took a couple of hours per model, which totals around 100 hours of GPU runtime for all experiments. 
We use OpenAI's \texttt{GPT-4.1} as a comparison point, prompted in a one-shot manner. We report all the prompts used for all tasks in appendix \ref{app:prompts}. Figures \ref{fig:F1_au_extraction}, \ref{fig:F1_as_extraction} and \ref{fig:bertscore_au_clarif} display the results of base, instruct and finetuned SLMs on the three tasks of AU extraction, AS detection and AU clarification compared to the \texttt{GPT-4.1} baseline.

The selected models for each task are \texttt{Qwen-7B}  ($lr=5e-5$) for AU extraction, \texttt{Gemma-9B} ($lr=5e-5)$ for AS Detection and \texttt{Gemma-9B} ($lr=3e-5$) for AU Clarification.

\begin{figure*}
    \centering
    \begin{subfigure}{0.49\textwidth}
        \centering
        \includegraphics[width=\linewidth]{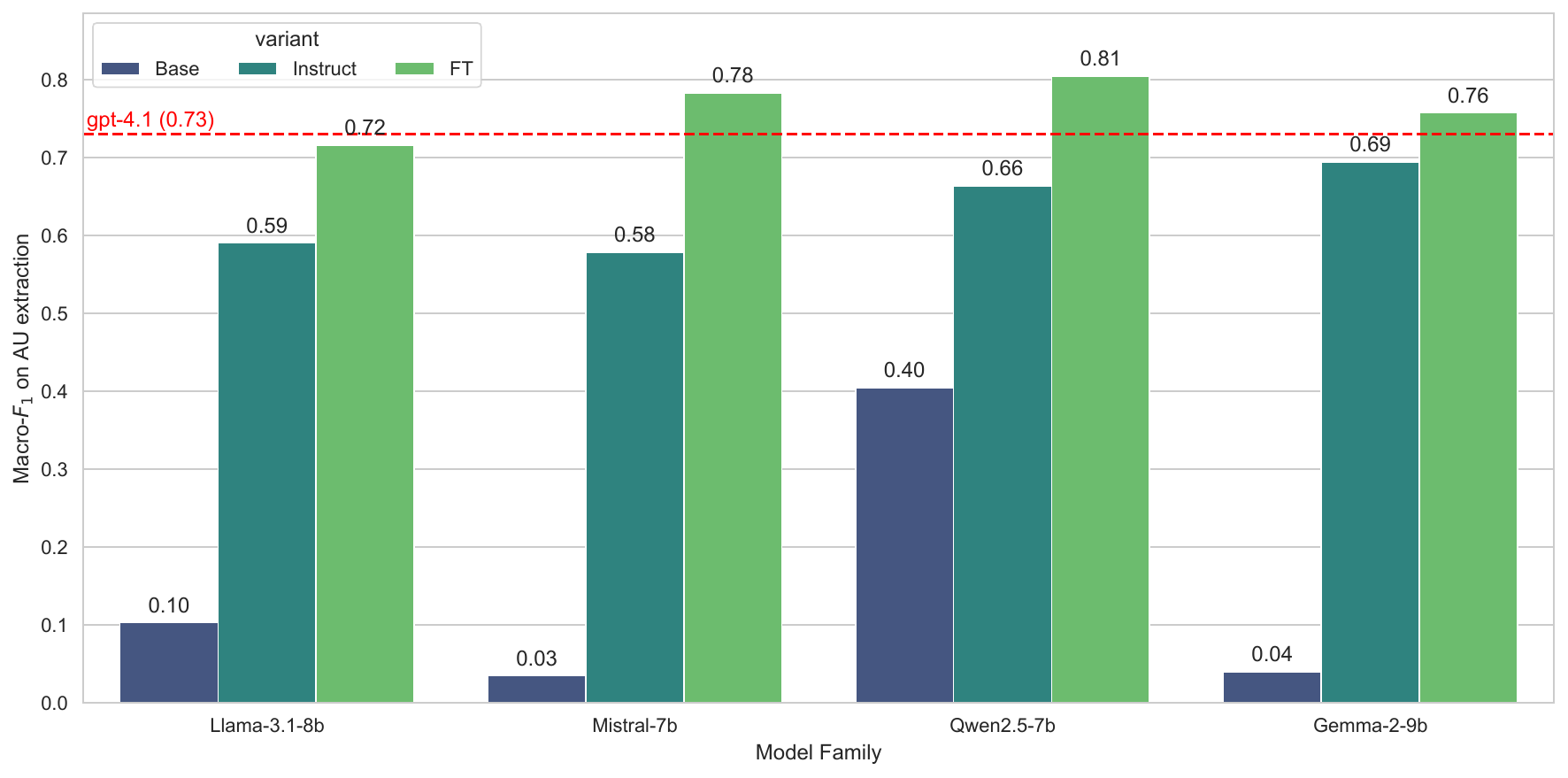}
        \caption{Macro-F1}
    \end{subfigure}
    \begin{subfigure}{0.49\textwidth}
        \centering
        \includegraphics[width=\linewidth]{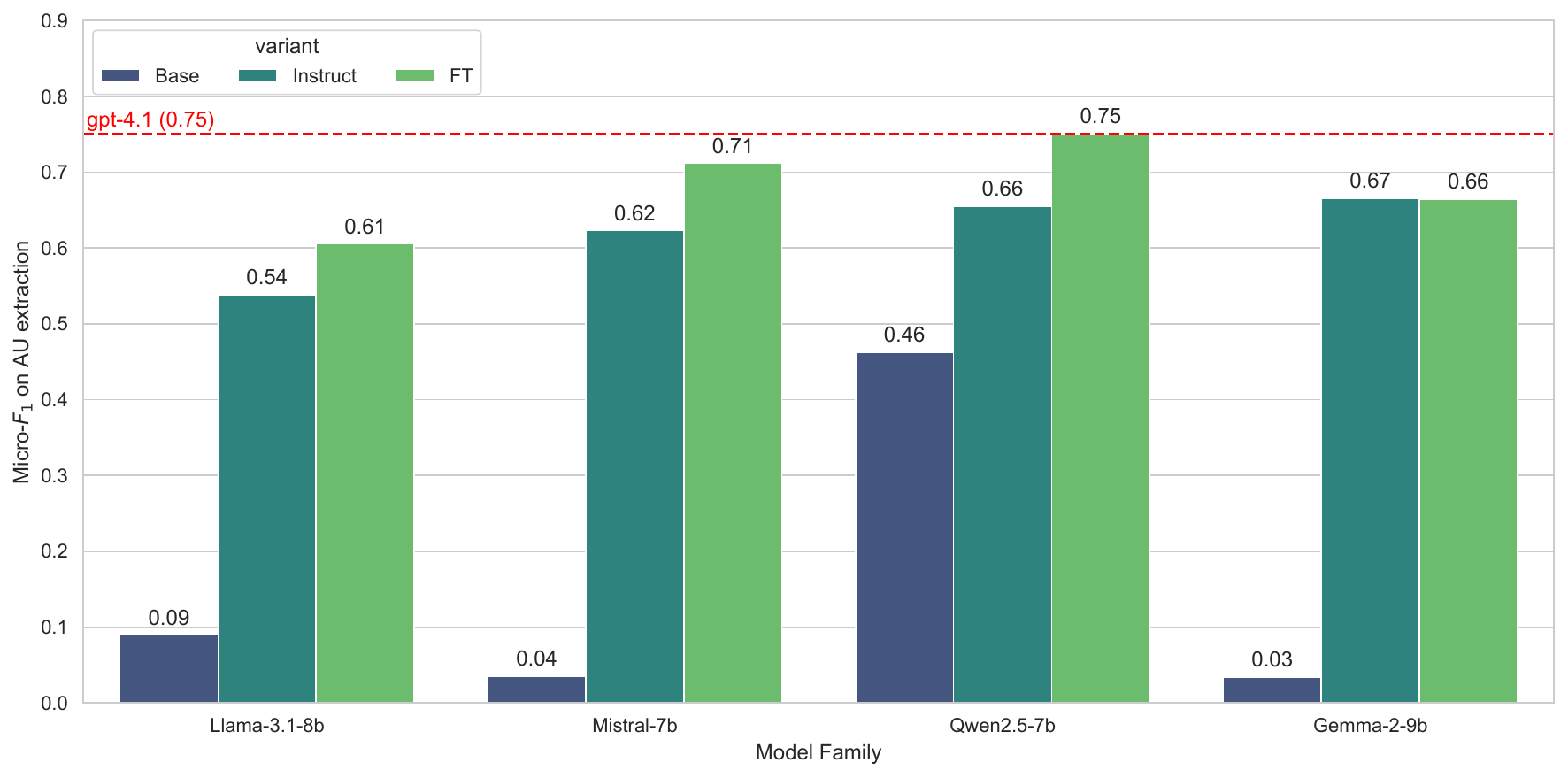}
        \caption{Micro-F1}
    \end{subfigure}
    \caption{Macro and Micro F1 of different model families on the task of Argumentative Units extraction}
    \label{fig:F1_au_extraction}
\end{figure*}

\subsection{Encoder-based Approaches}
\label{subapp:encoder}

\subsubsection{Argument Unit Detection}
\label{subsubapp:encoder_AU}

\textit{Argumentative Unit Detection} amounts to identifying content containing argumentative structure in texts, which can be performed using encoders in several ways. We experimented with two encoder-based formulations for argumentative unit (AU) extraction: a BIO token classifier where every token is labeled as outside (O), inside (I) or the beginning of a relevant sequence and an enumerative span classifier, which scores all possible spans up to a maximum length. Overall, BIO models performed better.

All models are trained with AdamW, a linear warmup scheduler, gradient clipping, and early stopping on validation micro-F1. We convert token index predictions into textual spans to use the same evaluation functions as those used for decoder-based segmentations.

\paragraph{BIO token classification}

Our main model casts AU extraction as sequence labeling with three labels: \texttt{O}, \texttt{B-AU}, and \texttt{I-AU}. Given an input sequence $x=(x_1,\dots,x_T)$, the encoder produces contextualized token representations $h_1,\dots,h_T$, and a linear classifier predicts a label distribution for each token:
\[
p(y_t \mid x) = \mathrm{softmax}(W h_t + b).
\]

Gold spans are projected onto token indices: the first token of a span receives \texttt{B-AU}, following tokens receive \texttt{I-AU}, and all others receive \texttt{O}. Special tokens and padding are ignored in the loss. Training uses standard token-level cross-entropy. At inference time, predicted BIO sequences are converted back into spans using tokenizer offsets.

\paragraph{Span classification}

We also implemented a span classifier that scores candidate spans directly. Given an input sequence of tokens $(t_1, \dots, t_n)$, the encoder produces contextualized
representations $(\mathbf{h}_1, \dots, \mathbf{h}_n)$. For each candidate segment $(i, j)$ with $1 \leq i \leq j \leq n$, we construct a segment
representation by concatenating the representation of the start token $\mathbf{h}_i$, the representation of the end token $\mathbf{h}_j$, the average representation of tokens within the span $\frac{1}{j-i+1} \sum_{k=i}^{j} \mathbf{h}_k$, and the representation of the special \texttt{[CLS]} token $\mathbf{h}_0$. This representation is passed to a small feed-forward classifier producing a scalar span score. Training uses binary cross-entropy over candidate spans, with positive-class reweighting to address class imbalance.

At inference time, spans are decoded greedily from left to right using a score threshold. Although this formulation predicts spans directly, it is computationally more expensive and performed worse than BIO in our experiments.

\paragraph{Model selection}

We tuned encoder backbone, learning rate, batch size, and dropout for both models, and additionally varied maximum span length for the span classifier. The BIO formulation gave the best validation performance, so we retain it as our main encoder-based AU extraction model. Best results are displayed in Table~\ref{tab:AU_span_eval}.

\begin{table}[t]
\centering
\setlength{\tabcolsep}{6pt}
\small
\begin{tabular}{lc}
\toprule
\textbf{Hyperparameter} & \textbf{Value} \\
\midrule
Model & \texttt{almanach/camembertav2-base} \\
Learning rate & $2 \times 10^{-5}$ \\
Dropout & 0.1 \\
Batch size & 4 \\
\bottomrule
\end{tabular}
\caption{\centering Best Hyperparameter Configuration for Encoder-based Argument Unit Detection}
\label{tab:AU_best_hparams}
\end{table}

\begin{table}[h!]
\centering
\setlength{\tabcolsep}{5pt}
\normalsize
\begin{tabular}{lccc}
\toprule
\textbf{Metric} & \textbf{Precision} & \textbf{Recall} & \textbf{F1} \\
\midrule
Macro & 0.879 & 0.824 & 0.831 \\
Micro & 0.869 & 0.782 & 0.823 \\
\bottomrule
\end{tabular}
\caption{\centering Macro and Micro F1-scores on the Test Set for Encoder-based Argument Unit Detection ($\lambda_{\text{overlap}} = 0.5$)}
\label{tab:AU_span_eval}
\end{table}

\subsubsection{Argument Structure Detection}
\label{subsubapp:encoder_AS}
As \textit{Argumentative Structure Detection} can be formulated as a span extraction and tagging task, we explore the capacities of an encoder-based model for this task, which is faster and more memory-efficient to train and use. This type of models could be useful in resource-constrained scenarios, and prove to be reliable for this task. 


\paragraph{Model Architecture} We use a span representation similar to that of the span-based encoder approach for AU detection. For each candidate span $(s,e)$ up to a maximum length, we build a representation by concatenating the start token, end token, mean-pooled span representation, and CLS representation:
\[
r_{s,e} = [h_s ; h_e ; \bar{h}_{s:e} ; h_{\mathrm{0}}].
\]
This results in a fixed-size representation for each candidate span, independently of its length.

\paragraph{Training Objective}
The model is trained to jointly predict the end position and the type of each gold span. For each gold segment $(s_k, e_k, \tau_k)$ in the training set, the start token is fixed to its gold value $s_k$ (the beginning of the gold span), and we enumerate all candidate ends, thereby considering all possible spans up to a maximum span length, compute their representations and their scores using a feedforward scoring head. We use a cross-entropy loss over candidate ends with the gold end token $e_k$ as reference. Similarly, a cross-entropy loss is performed over types using the gold type $\tau_k$. The total loss is defined as:
\[
\mathcal{L} = \mathcal{L}_{\text{end}} + \lambda \cdot \mathcal{L}_{\text{type}}
\]
where $\mathcal{L}_{\text{end}}$ is the average loss for end prediction, $\mathcal{L}_{\text{type}}$ is the average loss for type classification, and $\lambda$ is a tunable weight (\texttt{type\_loss\_weight}, set via hyperparameter optimization).

\paragraph{Inference Procedure}

Since each Argumentative Unit (AU) is assumed to contain only relevant argumentative material (as opposed to full-text detection), we adopt a greedy decoding strategy that sequentially consumes the entire sequence without leaving gaps. This design reflects the annotation convention of the corpus, in which AUs are extracted first and every token inside an AU belongs to exactly one argumentative span. Accordingly, we perform inference using a sequential greedy strategy to extract text spans and assign each one an argument type. Let $\mathbf{x} = \texttt{[CLS]}, x_1, x_2, \ldots, x_T, \texttt{[SEP]}$ be the tokenized input sequence, and $\mathbf{H} = [\mathbf{h}_0, \mathbf{h}_1, \ldots, \mathbf{h}_T]$ be the corresponding contextual representations obtained from the encoder, where $\mathbf{h}_0$ corresponds to the \texttt{[CLS]} token. Given a start token index $s$, we enumerate all possible spans $(s,e)$ up to the maximum possible length (or end of the whole sequence) and compute their scores for span and type as:
\[
    (\text{score}_{s,e}, \tau_{s,e}) = f_{\text{span}}(\mathbf{h}_s, \mathbf{h}_e, \mathbf{h}_{\text{mean}}, \mathbf{h}_0)
\]
where $f_{\text{span}}$ is a span scoring function and $\mathbf{h}_{\text{mean}}$ is the mean-pooled representation of all token hidden states from index $s$ to $e$. The score $\text{score}_{s,e}$ assesses the likelihood of the span being argumentative, and $\tau_{s,e}$ denotes its associated type. We select the best end point based on the span score:
\[
    (e^*, \tau_{s,e^*}) = \arg\max_{e} \; \text{score}_{s,e}
\]

\paragraph{Hyperparameter Tuning}
We optimize all hyperparameters with Optuna \citep{akiba2019optuna}. The search space includes the choice of encoder (\texttt{camembertav2-base} \citep{antoun2024camembert20smarterfrench} or \texttt{moderncamembert-base} \citep{antoun2025modernbertdebertav3examiningarchitecture}), learning rate (log-uniform), dropout rate, maximum span length, and the weighting of the type-classification loss. We use Optuna’s Median Pruner to discard low-performing configurations early based on development micro-F1. For each trial, models are trained with a batch size of 4 examples and using early stopping, for up to a maximum of 10 epochs. The final selected configuration yielded a micro-F1 score of 0.704 over the development set after 4 epochs, and is shown in Table~\ref{tab:AS_best_hparams}. 

\begin{table}[t]
\centering
\setlength{\tabcolsep}{6pt}
\small
\begin{tabular}{lc}
\toprule
\textbf{Hyperparameter} & \textbf{Value} \\
\midrule
Model & \texttt{almanach/camembertav2-base} \\
Learning rate & $5.379 \times 10^{-5}$ \\
Dropout & 0.202 \\
Type loss weight & 0.878 \\
Max span length & 261 \\
\bottomrule
\end{tabular}
\caption{\centering Best Hyperparameter Configuration for Encoder-based Argument Structure Detection selected via Optuna}
\label{tab:AS_best_hparams}
\end{table}

\paragraph{Encoder Performances}
Micro- and macro-metrics similar to those presented for decoder-based approaches are reported in Table~\ref{tab:AS_span_eval}.

\begin{table}[h!]
\centering
\setlength{\tabcolsep}{5pt}
\normalsize
\begin{tabular}{lccc}
\toprule
\textbf{Metric} & \textbf{Precision} & \textbf{Recall} & \textbf{F1} \\
\midrule
Macro & 0.769 & 0.779 & 0.765 \\
Micro & 0.697 & 0.711 & 0.704 \\
\bottomrule
\end{tabular}
\caption{\centering Macro and Micro F1-scores on the Test Set for Encoder-based Argument Structure Detection ($\lambda_{\text{overlap}} = 0.5$)}
\label{tab:AS_span_eval}
\end{table}

\section{Statistical Model for LLMs clarification qualities}
\label{app:opti}
In this appendix, we elaborate the statistical model introduced in Section~\ref{part:corpus}.

We formalize the AI-human hybrid annotation process as a decision model to evaluate the intrinsic quality of the four clarification systems. Following the argumentative segmentation, an annotator requests an automated clarification from a model $l$, where $l$ is sampled from a discrete uniform distribution $L \sim \mathcal{U}\{1, \dots, 4\}$. The annotator then makes a binary decision based on an internal quality threshold $\tau_k$, which varies depending on the number of attempts $k$. 
If the generated clarification quality $e$ meets or exceeds the threshold ($e \ge \tau_k$), the annotator accepts the output. In this case, we directly observe the quality $e$ as the ROUGE-L score between the model's output and the final human-validated text. Else, if the quality falls below the threshold ($e < \tau_k$), the annotator rejects the output and requests a new generation, leaving the specific value of $e$ unknown. We model the quality $e$ for each model $l$ using a Beta distribution, $e \sim Beta(\alpha_l, \beta_l)$. Observations are $\mathcal{D} = \{(k,l,r,e)_j\}_{j \in data}$ where $r \in \{0, 1\}$ is the acceptance indicator and $\tau_k$ is modeled as a Dirac distribution dependent on the attempt index $k$.
The total log-likelihood for our observations $\mathcal{D}$ is defined as follows:


\begin{equation}
\begin{split}
\log\mathcal{L} &= \sum \log p(r_j,e_j|k_j,l_j) \\ 
&= \sum_{j | r=0} \log p(e_j | k_j, l_j) \cdot p(e > \tau_{k_j} | k_j, l_j)  \\& + \sum_{j|r=1}\log p(e < \tau_{k_j}|k_j,l_j) \\
\end{split}
\end{equation}

We jointly optimize the parameters $\alpha_l$ and $\beta_l$ and the acceptance thresholds $\tau_k$ via gradient ascent to find the maximum likelihood estimates for each system.

As the Beta distribution's cumulative distribution function (defined as the regularized incomplete beta function $I_{x}(a,b)={\frac {\mathrm {B} (x;\,a,b)}{\mathrm {B} (a,b)}}$)
does not have a closed form and its partial derivatives are not tractable, we approximate its value through numerical integration and compute the gradients by implementing it with torch.

The final $\alpha_l$ and $\beta_l$ after optimization are reported in Table \ref{tab:alpha_beta}. The associated means computed as $\frac{\alpha_l}{\alpha_l + \beta_l}$ are reported in Section \ref{part:annotation}.

\begin{table}
  \centering
  \begin{tabular}{l|cc}
    \hline
     & $\alpha $ & $\beta$ \\
    \hline
    \texttt{GPT-4.1} & 1.75 & 1.12e-1\\
    \texttt{Llama-70B} & 1.80 & 1.22e-1\\
    \texttt{Qwen-32B} & 1.57 & 1.19e-1\\
    \texttt{Llama-8B} & 1.10 & 1.16e-1\\
    \hline
  \end{tabular}
  \caption{
    $\alpha_l$ and $\beta_l$ for all four models.
  }
   \label{tab:alpha_beta}
\end{table}








\section{Clarification Improvements after finetuning}
\label{app:llmasjudge_clarifs}
Figure \ref{fig:prompt_llmasjudge_clarifs} shows the prompt used for LLM-as-a-Judge for the clarification evaluation. Over the 348 examples of the test set, the preferred clarification was the one of the LLM during the annotation 91 of the time, the one of the finetuned model 231 times, and a draw 26 times.   

We use the statistical test $\chi^2$, where $H_0$ is the fact that all three possible outcomes are equally likely, and the binomial statistical tests by eliminating the draw option and considering $H_0$ as A and B are equally likely.
For both tests, the null hypothesis $H_0$ can be rejected ($p<0.001$).

Interestingly, by doing the same evaluation by comparing the finetuned SLMs and annotators' clarifications, the finetuned SLMs are preferred in 194 examples (111 for annotators, 43 draws). this result is also statistically significant ($p<0.001$). We find that the finetuned SLM tends to rephrase the argumentative unit texts less than the LLMs, and therefore of annotators clarifications (which are usually modifications of LLMs output, as highlighted in appendix \ref{app:corpus_analysis}). This makes their clarifications closer to the original meaning of the AUs while correcting their writing mistakes and unnatural phrasings. Table \ref{tab:examples_clarifs} provides examples of SLMs output that avoid the over-analysis of the LLM model.

\begin{figure*}[ht]
  \begin{promptbox}
  \begin{verbatim}
### Role
You are an expert in rewriting and clarification. Your task is to judge the 
clarity of a given text.

### Strict instructions
The user will give you a text, a segment of that text (which may 
potentially be the entire text), and two clarifications, A and B, of that segment.
A clarification must transform the initial text into a clear and self-sufficient 
text that can be understood without the context of the initial text.
This clarification may add this context if it is contained in the initial text
and rephrase the segment. However, it must not add
justifications if the text does not mention them. You must judge which of the two
clarifications is the best.
Answer ONLY with “A,” “B,” or “TIE.” Answer “TIE” only if there is 
no difference between the two. Prefer “A” or “B.” 

### Reference examples:
Example 1 (adding justification):
Text: 
I don't understand how anyone can be so out of touch with reality. THE PRESIDENT'S
SALARY MUST BE REDUCED!!
Segment: 
THE PRESIDENT'S SALARY MUST BE REDUCED!!
Clarifications:
A- The president's salary must be lowered.
B- The president's salary must be lowered to limit public spending.
Answer: A

Example 2 (one of the two adds important context):
Text: 
Inequalities are too great; aid must be increased. The same goes for tax audits.
Segment:
Segment: 
The same applies to tax control.
Clarification:
A- Tax control must be strengthened. 
B- Tax control must be strengthened because inequality is too high.
Answer: B

Example 3 (equality):
Text: 
Lower the allowances of parliamentarians.
Segment: 
Lower the allowances of parliamentarians.
Clarification:
- The allowances of parliamentarians must be lowered.
- Lower the allowances of parliamentarians.
Answer: EQUALITY
\end{verbatim}
  \end{promptbox}
  \caption{LLM-as-a-judge prompt for clarification evaluation (translated from French to English).}
  \label{fig:prompt_llmasjudge_clarifs}
\end{figure*}

\begin{table*}
  \centering
  \begin{tabularx}{\textwidth}{ X|X|X }
    \hline
    \textbf{Initial AU}  & \textbf{LLM output} & \textbf{Finetuned SLM output} \\
    \hline
    Mettre en place des centres de collecte (vêtements matériels puériculture jouets) pour les enfants malades ou placés. Idem pour les animaux. Idem pour le recyclage.
    \newline
    \newline
    \small \textit{Set up collection centers (clothing, childcare items, toys) for sick or foster children. The same applies to animals. The same applies to recycling.} & Mettre en place des centres de collecte pour vêtements, matériel de puériculture et jouets destinés aux enfants malades ou placés, ainsi que pour les animaux et le recyclage, permettrait d’apporter une aide concrète et ciblée en complément de la fiscalité classique.
    \newline
    \newline
    \small \textit{Setting up collection centers for clothing, childcare equipment, and toys for sick or foster children, as well as for animals and recycling, would provide concrete and targeted assistance in addition to traditional taxation.} & Il faut mettre en place des centres de collecte pour les enfants malades ou placés, les animaux et le recyclage.
    \newline
    \newline
    \small \textit{Collection centers must be set up for sick or foster children, animals, and recycling.}\\
    \hline
    Taxer les propriétaires des grosses cylindrées.
    \newline \newline
    \small \textit{Tax owners of large-engine vehicles. } 
    & Il faut taxer les propriétaires des grosses cylindrées pour encourager des choix de véhicules moins polluants.'
    \newline \newline
    \small \textit{Owners of large-engine vehicles should be taxed to encourage the choice of less polluting vehicles.} 
    & Taxer les propriétaires des grosses cylindrées. 
    \newline \newline
    \small \textit{Tax owners of large-engine vehicles. }\\
    \hline
    Subventions état pour stérilisation chats errants . \newline \newline 
    \small \textit{Government subsidies for sterilization stray cats .}
    & L'État devrait accorder des subventions pour la stérilisation des chats errants afin de mieux protéger les animaux. \newline \newline
    \small \textit{The government should provide subsidies for the sterilization of stray cats in order to better protect animals.} & Il faut accorder des subventions publiques pour la stérilisation des chats errants. \newline \newline
    \small \textit{Public subsidies should be granted for the sterilization of stray cats.}\\
    \hline
  \end{tabularx}
  \caption{
    Some example clarifications from the test set given by the LLMs during the annotation and by the finetuned SLM after finetuning.      
  }
  \label{tab:examples_clarifs}
\end{table*}

\section{Clustering Improvements}
\label{app:clustering}
\paragraph{Technical details} We use the \texttt{paraphrase-multilingual-MiniLM-L12-v2} sentence transformer model \cite{reimers-2019-sentence-bert} as embedding model, UMAP \cite{sainburg2021parametric} as a dimension reduction algorithm and HDBSCAN \cite{campello2013density} as the clustering method. 

\paragraph{Experiments and results} Table \ref{tab:cluster_results} displays the results of the LLM-as-a-Judge. the lines Tax., Eco., Org. and Dem. respectively correspond to the themes \textit{Taxation and Public Spending}, \textit{Ecological Transition}, \textit{Organization of The State} and \textit{Democracy and Citizenship}. The three columns are the three experimental  settings: 
\begin{itemize}[itemsep=0pt, parsep=1pt, partopsep=1pt, topsep=1pt]
    \item \textbf{Setting 1}: Initial contributions (A) vs. Argumentative units (B)
    \item \textbf{Setting 2}: Argumentative units (A) vs. Clarifications (B)
    \item \textbf{Setting 3}: Argumentative units (A) vs. AU using clarification-based clusters (B)
\end{itemize}
    
in all three settings, D means \textit{Draw}. We evaluate the $\chi^2$ and binomial statistical tests with the same null hypothesis $H_0$ as done in Appendix \ref{app:llmasjudge_clarifs}.
For both tests, in all settings, and for all themes the null hypothesis $H_0$ can be rejected ($p<0.001$). 

\begin{table}[H]
\centering
\small 
\begin{tabular}{@{} l ccc ccc ccc @{}}
\toprule
\multirow{2}{*}{} & \multicolumn{3}{c}{\small \textbf{Setting 1}} & \multicolumn{3}{c}{\textbf{Setting 2}} & \multicolumn{3}{c}{\textbf{Setting 3}} \\
\cmidrule(lr){2-4} \cmidrule(lr){5-7} \cmidrule(lr){8-10}
 &\small A &\small B &\small D &\small A &\small B &\small D &\small A &\small B &\small D \\
\midrule
\small Tax & 15 & 81 & 4 & 6 & 92 & 2 & 13 & 87 & 0 \\
\small Eco & 22 & 77 & 1 & 14 & 84 & 2 & 8 & 91 & 1 \\
\small Org & 7 & 90 & 3 & 6 & 90 & 4 & 1 & 95 & 4 \\
\small Dem & 10 & 88 & 2 & 1 & 98 & 1 & 4 & 88 & 8 \\
\midrule
\small \textbf{Tot.} & \textbf{54} & \textbf{336} & \textbf{10} & \textbf{27} & \textbf{364} & \textbf{9} & \textbf{26} & \textbf{361} & \textbf{13} \\
\bottomrule
\end{tabular}
\caption{LLM-as-a-Judge preference distribution across four themes and three experimental settings.}
\label{tab:cluster_results}
\end{table}

Because the text type (segmented AUs) evaluated by the LLM-Judge was identical in Setting 3, the 90\% preference confirms that the improvement stems entirely from better cluster organization, not just linguistic clarity.

\section{Annotation and Finetuning Prompts}
\label{app:prompts}
 We display in this appendix the prompts used during the annotation (Figure \ref{fig:prompt_annot_clarif}), for the Argumentative Unit extraction task (Figure \ref{fig:prompt_AU_extraction}), the Argumentative Structure detection task (Figure \ref{fig:prompt_AS_detection}) and the Argumentative Unit extraction task (Figure \ref{fig:prompt_AU_clarif}). Figure \ref{fig:prompt_llm_as_judge} displays the LLM-as-a-Judge prompt used for the evaluation of clustering. 
 
All prompts were given in French to AI models in order to limit code switching, but are here given in English. For \texttt{GPT-4.1} prompting as a comparison baseline, we added one example in each prompt before giving the actual contribution to process.

\begin{figure*}[ht]
  \begin{promptbox}
  \begin{verbatim}
You are an argument clarification portal. The user will give you a written opinion 
on a given topic, as well as the segmentation of one of the arguments in that 
opinion into three types of segments: statement(s), argument(s), and solution(s).
Extract, in one sentence, the clear and self-sufficient argument underlying this 
segmentation. Prioritize the solution, and include arguments and statements 
only if they seem relevant to you. You can use the context of the entire opinion 
to help you, but do not include any information that is not present 
in the segments. Respond only with the clear and self-sufficient argument, and 
nothing  else. If the argument is already clear and well written, you can refer 
directly to that argument.

Given the opinion: 
{{contribution}}

on the topic {{theme}}

Extract, in one sentence, the underlying argument consisting of:
- Statements: {{statements}}
- Arguments: {{premises}}
- Solutions: {{solutions}}
\end{verbatim}
  \end{promptbox}
  \caption{System Prompt for the clarification task used during the annotation (translated from French to English).}
  \label{fig:prompt_annot_clarif}
\end{figure*}

\begin{figure*}[ht]
  \begin{promptbox}
  \begin{verbatim}
I am going to give you an opinion piece in French. Your task is to segment this 
text into argumentative units. We define an argumentative unit as one
or more segments of the text that focus on a particular topic. It may consist of 
solutions, arguments, or simple statements. An argumentative unit 
is not necessarily contiguous: it can join segments that do not follow each other.

This task is EXTRACTIVE. You must COPY and only copy the text of the argumentative 
unit exactly as it is written, including capital letters and punctuation. 
If the argumentative unit is composed of several non-contiguous segments,
 you can concatenate them by simply separating them with a space. There is at 
least one argumentative unit in the text, but no maximum number. Highlight the 
argumentative units in the form of a list as shown in the example. 
Not all segments of the text are necessarily part of an argumentative unit.

You must give the argumentative units in the form of a list:  
- argumentative unit 1
- argumentative unit 2...

Do NOT output ANYTHING OTHER than the list of argumentative units.
Here is the text:
{{contribution}}
\end{verbatim}
  \end{promptbox}
  \caption{System Prompt for the Argumentative Units extraction tasks (translated from French to English). }
  \label{fig:prompt_AU_extraction}
\end{figure*}

\begin{figure*}[ht]
  \begin{promptbox}
  \begin{verbatim}
I am going to give you a segment of text containing opinions in French. Your task is 
to segment this text and assign each segment a type. The possible types are 
STATEMENT, PREMISE, and SOLUTION, and ONLY those. Below is the definition of 
each type:
- SOLUTION: a proposal for action (whether concrete and feasible or not) to be 
taken to solve a problem.
- STATEMENT: the expression of an opinion as an assertion, which does not provide a 
solution but rather expresses a feeling.
- PREMISE: a justification, argument, or example that supports an 
assertion or solution.

This task is EXTRACTIVE; you must copy the text of each segment exactly 
as it is written, including capital letters and punctuation.
The entire text must be segmented. Not all types of 
segments are necessarily present, and several segments may be of the same type.
You MUST highlight the segmentation by following the exact format of the example, 
including the “-” for each segment.

- [STATEMENT] Statement 1
- [SOLUTION] Solution 1
- [STATEMENT] Statement 2
- [PREMISE] Argument 1...

I will give you the original text and the segment, and you must output the list of 
segments and their types in the form “- [TYPE] SEGMENT,” and nothing else.

Original text:
{{contribution}}

Segment:
{{argumentative unit}}
\end{verbatim}
  \end{promptbox}
  \caption{System Prompt for the Argumentative Structure detection task (translated from French to English).}
  \label{fig:prompt_AS_detection}
\end{figure*}

\begin{figure*}[ht]
  \begin{promptbox}
  \begin{verbatim}
I will give you a segment of opinion text in French, as well as the original text 
from which it is taken. Your job is to rewrite this opinion segment 
clearly. In particular, you must first correct spelling, 
grammar, and syntax errors. You must also add the context present in the original 
text  if it is important for understanding the segment. The final 
clarified text must be understandable without access to either the original text or 
its sub-segment. If the segment is already clear and well written, you should 
simply copy it. You should ONLY highlight the clarification of the segment and 
nothing else.

Original text:
{{contribution}}

Segment to be clarified:
{{argumentative unit}}
\end{verbatim}
  \end{promptbox}
  \caption{System Prompt for the Argumentative Unit clarification task (translated from French to English.}
  \label{fig:prompt_AU_clarif}
\end{figure*}

\begin{figure*}[ht]
  \begin{promptbox}
  \begin{verbatim}
### Role
You are an expert judge of thematic consistency. Your task is to compare two 
clusters of texts (A and B).

### Strict instructions
1. A group is “better” if it is more specific, more precise, and if ALL 
texts deal with the same subject using the same approach.
2. If a group contains different subjects (even if they are vaguely related to 
politics or money), it must be penalized.
3. Respond ONLY with “A,” “B,” or “TIE.” Respond with “TIE” only if there is 
no difference between the two. Give preference to “A” or “B.” 
No explanations will be tolerated.

### Reference examples:
Example 1 (Divergent topics vs. Identical topics):
A: 
- “Reform public inquiries”
- “Limit aid to foreigners”
- “Climate change”
B: 
- “Tax kerosene,” 
- “Tax polluting companies”
Verdict: B

Example 2 (Similar but different topics vs. Identical topics):
A: 
- “Inheritance and estate taxes,” 
- “Cost of retirement homes/nursing homes”
B: 
- “Eliminate compensation for former presidents,” 
- “Reduce benefits for former presidents”
Verdict: B

Example 3 (Total consistency on both sides):
A: 
- “Legalize cannabis”
- “Sell cannabis in pharmacies”
B: 
- “Increase the minimum wage,” 
- “Raise the minimum wage”
Verdict: TIE
\end{verbatim}
  \end{promptbox}
  \caption{System Prompt for the LLM-as-a-Judge evaluation of clustering (translated from French to English).}
  \label{fig:prompt_llm_as_judge}
\end{figure*}
    

%% file: parts/summary_corrections.tex




\subsection{Characterization of Annotators Corrections}
\label{subapp:caraterization_corrections}
In this section, we explore the corrections made by the annotators in details by computing the differences between the Argumentative Units, the clarification generated by AI systems, and the final clarifications accepted by the annotators using different metrics and systems.   

\paragraph{Levenshtein edit distance}
To better understand the modifications introduced by human annotators, we computed the edit distance between: (i) the original argumentative unit and the LLM clarification, (ii) the original argumentative unit and the final clarification, and (iii) the LLM clarification and the final clarification.

The edit distance, also known as Levenshtein distance, measures the similarity between two strings by counting the minimum number of character-level insertions, deletions, and substitutions required to transform one string into the other. The results are presented in Table \ref{tab:results_ed_distance}.

\begin{table*}
\centering
\begin{tabular}{l|cccc|c}
\hline
& \textbf{GPT-4.1} & \textbf{qwen3-32b} & \textbf{llama3.3-70b} & \textbf{llama3.1-8b} & \textbf{Mean} \\
\hline
\textbf{AU $\xrightarrow{}$ LLM} & 125 & 135 & 134 & 119 & 128 \\
\textbf{AU $\xrightarrow{}$ Clarif.} & 81 & 88 & 93 & 85 & 87 \\
\textbf{LLM $\xrightarrow{}$ Clarif.} & 78 & 81 & 71 & 74 & 76 \\
\hline
\hline
\textbf{AU $\xrightarrow{}$ LLM=Clarif.} & 134 & 166 & 139 & 141 & 145 \\
\hline
\end{tabular}
\caption{edit distance between the original argumentative unit (UA), the LLM clarification (LLM), and the final clarification (Clarif.). The last row corresponds to the cases for which the LLM output was accepted as-is by the annotator.}
\label{tab:results_ed_distance}
\end{table*}

On average, the edit distance between the original text and the LLM clarification is higher than between the original text and the final clarification. This suggests that LLMs tend to introduce additional information that is later removed by human annotators. This interpretation is consistent with the smaller edit distance observed between the LLM clarification and the final clarification.

Furthermore, 40\% of the final clarifications (331 out of 828) are entirely contained within the corresponding LLM clarification, ignoring punctuation. This indicates that, in many cases, the annotator’s intervention consists mainly in deleting superfluous content rather than rewriting the clarification. We further compare these results with the edit distances computed for argumentative units whose clarifications were not modified by annotators, shown in the last row of Table \ref{tab:results_ed_distance}. The edit distances in this case are relatively high and comparable to those observed between the original units and the LLM clarifications in Table \ref{tab:results_ed_distance}. This suggests that LLMs systematically attempt to reformulate the original text, even when no modification is ultimately necessary. When such changes are appropriate, the clarification is kept; otherwise, it is corrected. This observation is supported by length statistics. Argumentative units with modified clarifications are shorter on average (143 characters) than those with unmodified clarifications (203 characters). However, the LLM clarifications have a similar average length in both cases (178 characters). The final clarifications, after human correction, are significantly shorter, with an average length of 120 characters, reinforcing the hypothesis that human corrections primarily involve removing unnecessary content.


\paragraph{ROUGE scores}
ROUGE (Recall-Oriented Understudy for Gisting Evaluation) scores measure textual similarity by comparing overlapping units such as n-grams and longest common subsequences, typically between an automatically generated summary and a human reference summary. ROUGE scores range from 0 to 1, with higher values indicating greater similarity.

We report ROUGE-1 (unigram overlap), ROUGE-2 (bigram overlap), and ROUGE-L (longest common subsequence) scores. The results are shown in Table \ref{tab:rouge-scores}. Once again, the final clarifications appear more similar to the original argumentative units than the LLM clarifications.

\begin{table*}
\centering
\begin{tabular}{l|ccc}
\hline
& \textbf{ROUGE-1} & \textbf{ROUGE-2} & \textbf{ROUGE-L} \\
\hline
\textbf{LLM $\leftrightarrow$ Clarif.} & 0.68 & 0.63 & 0.67 \\
\textbf{LLM $\leftrightarrow$ AU} & 0.53 & 0.40 & 0.46 \\
\textbf{Clarif. $\leftrightarrow$ AU} & 0.62 & 0.48 & 0.56 \\
\hline
\hline
\textbf{LLM=Clarif. $\leftrightarrow$ AU} & 0.57 & 0.42 & 0.47 \\
\hline
\end{tabular}
\caption{ROUGE scores between the LLM clarification (LLM), the final clarification (Clarif.), and the original argumentative unit (AU). The last row corresponds to the cases for which the LLM output was accepted as-is by the annotator.}
\label{tab:rouge-scores}
\end{table*}

For argumentative units with unmodified clarifications, ROUGE scores between the clarification and the original unit (last row of Table \ref{tab:rouge-scores}) are close to the LLM-original scores observed above, which is consistent with our previous findings.


\paragraph{Perplexity}
Perplexity measures the uncertainty of a language model when predicting the next token in a sequence. We computed the perplexity of four LLMs (\texttt{gemma-2-9b-it}, \texttt{Meta-Llama-3.1-8B-Instruct}, \texttt{Mistral-7B-Instruct-v0.3}, and \texttt{Qwen2.5-7B-Instruct}) on the original argumentative units, the LLM clarifications, and the final clarifications. Results are shown in Table \ref{tab:perplexity}.

Across all models, the lowest perplexity is observed for the LLM clarifications, which is expected since these texts are generated by LLMs.

\begin{table*}
\centering
\begin{tabular}{l|cccc}
\hline
& \textbf{gemma-9b} & \textbf{Llama-8B} & \textbf{Mistral-7B} & \textbf{Qwen2.5-7B} \\
\hline
\textbf{AU} & 96.5 & 10.0 & 7.5 & 10.3 \\
\textbf{LLM clarification} & 41.9 & 4.6 & 4.0 & 4.9 \\
\textbf{final clarification} & 30.0 & 5.9 & 4.7 & 6.1 \\
\hline
\end{tabular}
\caption{Perplexity of the original argumentative unit, the LLM clarification, and the final clarification. We use the Instruct variant for all models.}
\label{tab:perplexity}
\end{table*}